\documentclass[lettersize,journal]{IEEEtran}
\usepackage{amsmath,amsfonts}
\usepackage{algorithmic}
\usepackage{algorithm}
\usepackage{array}
\usepackage[caption=false,font=normalsize,labelfont=sf,textfont=sf]{subfig}
\usepackage{textcomp}
\usepackage{stfloats}
\usepackage{url}
\usepackage{verbatim}
\usepackage{graphicx}
\usepackage{cite}
\usepackage{makecell}  
\usepackage{multirow}
\newcommand{\OurModel}[1]{CageAttack}
\newcommand{\firstpara}[1]{\noindent\textbf{{#1}.}~~}

\begin{document}
\title{Cage-Based Deformation for Transferable and Undefendable \\Point Cloud Attack}

\author{Keke Tang, Ziyong Du, Weilong Peng, Xiaofei Wang, Peican Zhu, Ligang Liu, and Zhihong Tian
\thanks{Keke Tang, Ziyong Du and Zhihong Tian are with the Cyberspace Institute of Advanced Technology, Guangzhou University, Guangzhou, Guangdong 510006, China.}
\thanks{Weilong Peng is with the School of Computer Science and Cyber Engineering, Guangzhou University, Guangzhou, Guangdong 510006, China.}
\thanks{Xiaofei Wang is with the  Department of Automation, University of Science and Technology of China,
Hefei, Anhui 230052, China.}
\thanks{Peican Zhu is with the School of Artificial Intelligence, Optics and
Electronics (iOPEN), Northwestern
Polytechnical University,
Xi’an, Shaanxi 710072, China.}
\thanks{Ligang Liu is with the Graphics \& Geometric Computing Laboratory, School
of Mathematical Sciences, University of Science and Technology of China,
Hefei, Anhui 230052, China.}
}

\markboth{Journal of \LaTeX\ Class Files,~Vol.~14, No.~8, August~2021}%
{Shell \MakeLowercase{\textit{et al.}}: A Sample Article Using IEEEtran.cls for IEEE Journals}


\maketitle

\begin{abstract}
Adversarial attacks on point clouds often impose strict geometric constraints to preserve plausibility; however, such constraints inherently limit transferability and undefendability. While deformation offers an alternative, existing unstructured approaches may introduce unnatural distortions, making adversarial point clouds conspicuous and undermining their plausibility. 
In this paper, we propose CageAttack, a cage-based deformation framework that produces natural adversarial point clouds. It  first constructs a cage around the target object, providing a structured basis for smooth, natural-looking  deformation.  
Perturbations are then applied to the cage vertices, which seamlessly propagate to the point cloud, ensuring that the resulting deformations remain intrinsic to the object and preserve plausibility.
Extensive experiments on seven 3D  deep neural network classifiers across three datasets show that CageAttack achieves a superior balance among transferability, undefendability, and plausibility, outperforming state-of-the-art methods. Codes will be made public upon acceptance.
\end{abstract}

\begin{IEEEkeywords}
Adversarial attacks, point clouds, cage-based deformation.
\end{IEEEkeywords}

\section{Introduction}

\IEEEPARstart{W}{ith} the rapid advancements in deep learning and depth-sensing technologies, deep neural networks (DNNs) have become the leading approach for 3D point cloud perception~\cite{guo2020deep,han2024mamba3d}. However, recent studies have shown that DNN classifiers are susceptible to adversarial attacks, where slight perturbations to input point clouds can cause incorrect predictions~\cite{Xiang-2019-3DAdversarialPCD,Liu-2019-extendingAdv3D}. This vulnerability poses significant challenges to deploying these systems in real-world scenarios. Therefore, investigating adversarial attacks on DNN classifiers for 3D point clouds is essential for assessing and enhancing their robustness against such threats~\cite{ijcai2021p591}.

{\em{A plausible adversarial point cloud looks genuine to human observers yet misleads a neural network.}} Typically, this plausibility is enforced  through strict imperceptibility constraints, requiring adversarial point clouds to remain nearly identical to the original by limiting point displacements~\cite{Xiang-2019-3DAdversarialPCD,Liu-2019-extendingAdv3D}. 
However, such constraints often hinder transferability across different models and make attacks less effective and even ineffective against adversarial defenses.
Although some methods attempt to mitigate these drawbacks via adversarial transformation models~\cite{liu2022imperceptible}, autoencoder-based reconstructions~\cite{Hamdi-2020-AdvPC}, or perturbation factorization~\cite{he2023-PF,chen2024anf}, 
their constraints on imperceptibility remain overly rigid, ultimately limiting transferability and undefendability.

\begin{figure}[!t]
\centering
\includegraphics[width=1\linewidth]{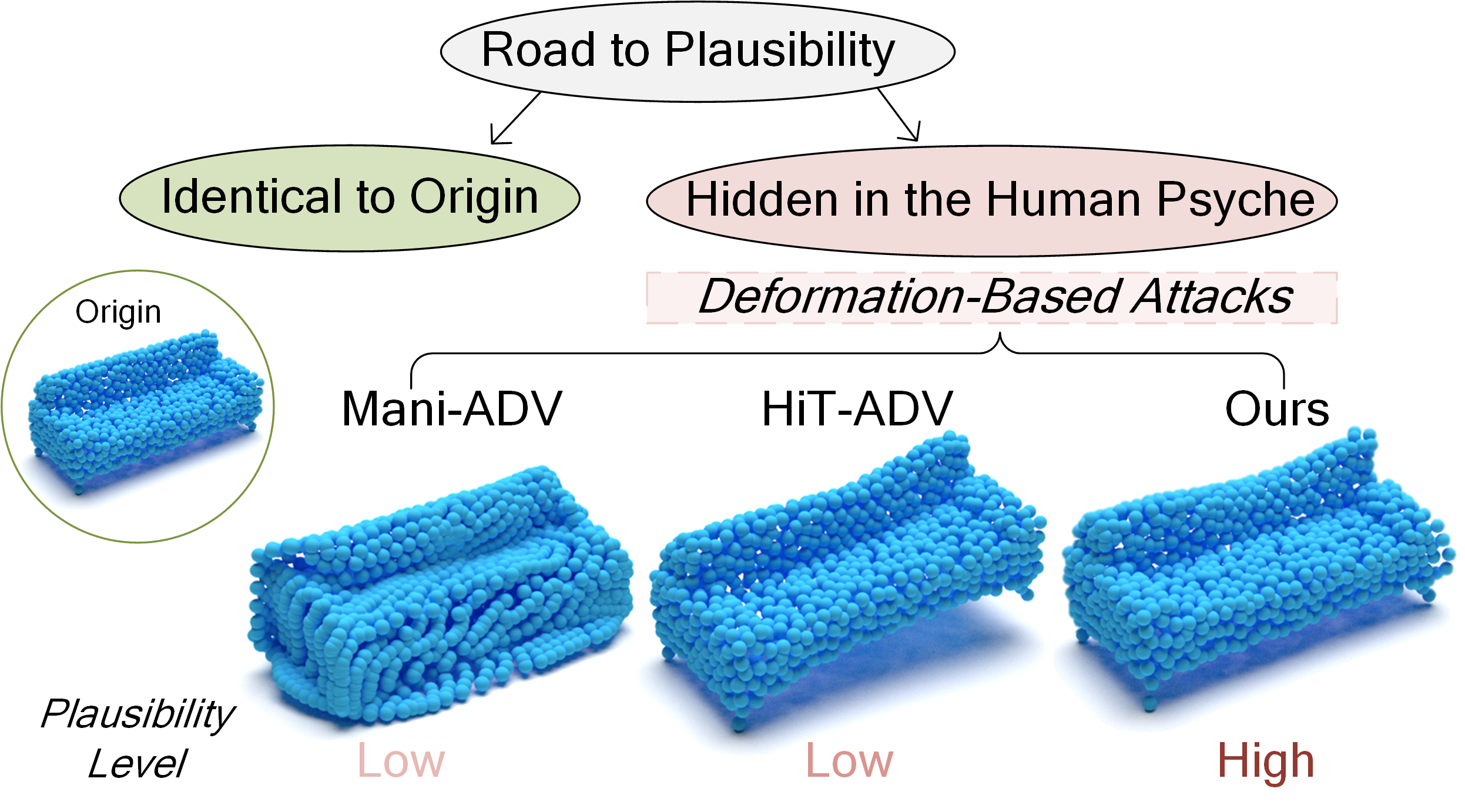} 
\caption{
Given a point cloud, adversarial attacks achieve plausibility by either enforcing strict geometric constraints to keep adversarial point clouds identical to the origin or leveraging natural deformations that remain \textit{hidden in the human psyche}. Existing deformation-based attacks, such as Mani-ADV~\cite{Tang-2023-ManifoldAttack} and HiT-ADV~\cite{lou2024hide}, often introduce unnatural distortions, making adversarial modifications conspicuous. In contrast, our method produces natural deformations, making changes intrinsic to the object and enhancing attack plausibility.
}
\label{fig:teaser}
\end{figure}

Therefore,
a more effective approach is to allow slight shape deformations rather than strictly preserving the original geometry.
As long as the deformation appears natural and intrinsic to the object, it remains \textit{hidden in the human psyche}~\cite{Xiang-2019-3DAdversarialPCD}, thereby achieving plausibility.
Mani-ADV~\cite{Tang-2023-ManifoldAttack} follows this principle by deforming the point cloud surface through the stretching of a parameter plane mapped to the object's shape. Similarly, HiT-ADV~\cite{lou2024hide} perturbs salient, imperceptible points and propagates the deformation to neighboring regions. 
While these methods improve transferability and undefendability, their non-structured deformation mechanisms often disrupt the naturalness of the shape, making adversarial point clouds more
conspicuous, and ultimately undermining plausibility, see Fig.~\ref{fig:teaser}.

To address the above issue, we propose \OurModel{}, a novel cage-based deformation framework for generating natural adversarial point clouds. Our approach begins by constructing a cage that encapsulates the target object, providing a compact representation of the point cloud and enabling global control over the deformation process. By perturbing the cage vertices, guided by gradients~\cite{Xiang-2019-3DAdversarialPCD}, these changes are smoothly propagated to the point cloud, resulting in natural deformations. This technique leverages the cage's structured nature to maintain shape consistency while applying controlled, subtle alterations. Compared to direct point-wise manipulation methods, cage-based modifications are 
more natural and thus unnoticeable.
We validate \OurModel{} by attacking seven widely used 3D DNN classifiers (e.g., the classic PointNet~\cite{Qi-2017-Pointnet} and the recent Mamba3D~\cite{han2024mamba3d}) across three datasets: synthetic ModelNet40~\cite{wu20153d}, ShapeNet Part~\cite{chang2015shapenet}, and the real-world ScanObjectNN~\cite{uy-scanobjectnn-iccv19}. Extensive experiments show that \OurModel{} achieves a superior balance of transferability,  undefendability, and plausibility, outperforming state-of-the-art methods.

\begin{itemize} 
\item 
We are the first to emphasize that preserving naturalness in deformation-based adversarial attacks on point clouds is crucial for maintaining plausibility.


\item We devise a novel deformation-based attack framework that utilizes the structured nature of the cage, enabling natural alterations that are less noticeable.

\item 
We show by experiments that our framework achieves a superior balance of transferability, undefendability, and plausibility compared to existing attack methods.
\end{itemize}

\section{Related Work}

\subsection{Adversarial Attacks on 3D Point Clouds}
Initially explored in the context of 2D image classification, adversarial attacks~\cite{chakraborty2021survey,ren2020adversarial,wei2024physical} have since been effectively adapted to 3D point cloud data. For 3D point clouds, these attacks are typically divided into three main categories: addition-based, deletion-based, and perturbation-based. Addition-based attacks introduce extra points to mislead classifiers~\cite{Xiang-2019-3DAdversarialPCD}, while deletion-based methods work by removing critical points to reduce classification accuracy~\cite{Zheng-2019-PCDSaliency, Yang-2019-AdvAttackAndDefense, Wicker-2019-IterSaliencyOcc, zhang-2021-TopologyDestructionNetwork}. Perturbation-based attacks, on the other hand, modify the positions of existing points to achieve adversarial effects~\cite{Xiang-2019-3DAdversarialPCD, zhao2020isometry, Kim-2021-minimalAdv,yang2025hiding}. This paper specifically focuses on perturbation-based approaches.

Early work by Xiang et al.~\cite{Xiang-2019-3DAdversarialPCD} and Liu et al.~\cite{Liu-2019-extendingAdv3D} adapted well-known 2D attack techniques such as C\&W~\cite{Carlini-2017-cw} and FGSM~\cite{Goodfellow-2014-FGSM} to 3D point clouds, laying the foundation for perturbation-based adversarial attacks in 3D. Since then, much research has focused on improving the imperceptibility of these attacks by preserving geometric and perceptual consistency. Representative strategies include preserving local curvature~\cite{wen2020geometry}, constraining perturbations along surface normals~\cite{liu2022imperceptible}, tangent directions~\cite{huang2022shape}, or their adaptive combination~\cite{Tang-NTAttack}, as well as leveraging manifold constraints~\cite{tang2024manifoldConstraints}, distributional uniformity~\cite{tang2024flat}, and structural symmetry~\cite{tang2024symattack}.

Beyond imperceptibility, recent efforts also target transferability and undefendability. AdvPC~\cite{Hamdi-2020-AdvPC}, for example, applies perturbations in the latent space before decoding, enabling adversarial examples to generalize better across different models. Building upon these developments, our approach relaxes overly strict imperceptibility constraints and instead allows for minor yet plausible shape deformations. This enables the generation of adversarial point clouds that maintain high perceptual quality while significantly improving transferability and resistance to defenses.

\subsection{Deformation-Based Point Cloud Attacks}

Deformation-based approaches introduce whole-shape deformations rather than applying perturbations to individual points, aiming to enhance transferability and undefendability. LGGAN~\cite{Zhou-2020-LGGAN} employs a generative adversarial network  to produce adversarial point clouds with slight deformations, enhancing both transferability and robustness against defenses. ShapeAdv~\cite{Lee-2020-Shapeadv} extends this approach by injecting perturbations directly into the latent space of an auto-encoder to generate transferable adversarial point clouds. Physical-aware methods, such as KNN-ADV~\cite{tsai2020robust}, adopt KNN distance constraints combined with a point-to-mesh reconstruction and resampling process to produce smoother deformed point clouds. Methods like MeshAttack~\cite{zhang20233d} and $\epsilon$-ISO~\cite{dong2022isometric} directly attack mesh data to ensure smooth deformations by employing edge length and Gaussian curvature regularizations, respectively.  Mani-ADV~\cite{Tang-2023-ManifoldAttack} applies perturbations within a parameter space for smoother deformations, while HiT-ADV~\cite{lou2024hide} achieves localized smoothness by applying Gaussian kernel-based deformations to specific regions.
In this paper, we propose a cage-based approach that induces more natural deformations to better balance  transferability,  undefendability, and plausibility.

\subsection{Deep  Point Cloud Classification}
Early DNN-based methods for  point cloud classification take voxel grids as inputs~\cite{Maturana-2015-voxnet}. The introduction of PointNet~\cite{Qi-2017-Pointnet} enabled raw point processing via multilayer perceptrons (MLP), leading to advanced MLP-based methods~\cite{Qi-2017-Pointnet++,PointMLP}, point-specific convolutions~\cite{Wu-2019-Pointconv,Thomas-2019-KPConv,Xu-2021-PAConv,Li-2018-PointCNN}, graph-based CNNs~\cite{Wang-2019-DGCNN,Zhao-2019-Pointweb,Shi-2020-PointGNN,chen2022ddgcn}, and recent architectures like Transformers and Mamba~\cite{zhao2021PT,guo2021pct,wu2022PT2,wu2024pT3,liang2024pointmamba,han2024mamba3d}. These advancements have significantly improved classification accuracy. 
For a comprehensive review, please refer to~\cite{ioannidou2017deep,guo2020deep}. 
This paper targets adversarial attacks on these classifiers.

\subsection{Geometric Deformation}  

Geometric deformation is a core task in computer graphics with a wide range of applications, e.g., shape animation~\cite{anguelov2005scape} and physics-based simulation~\cite{li2022diffcloth}. Key approaches include lattice-based methods, which provide precise local control over deformations~\cite{chen2011lattice}, and skeleton-based techniques, which are effective for realistic joint articulation~\cite{capell2002interactive, yoshizawa2007skeleton}. Cage-based deformation, which represents shapes using an enclosing cage structure, is particularly notable for its simplicity, efficiency, and ability to preserve naturalness with minimal distortion~\cite{nieto2012cage, garcia2013cages}. In this paper, we adopt cage-based deformation~\cite{stroter2024-CageSurvey} to generate natural-looking adversarial point clouds, that are transferable and undefendable.
\section{Problem Formulation}
\label{sec:problem}

\subsection{Problem Statement of Adversarial Attacks}

\firstpara{Typical Adversarial Attacks}
Given an object point cloud \( \mathcal{P} \in \mathbb{R}^{N \times 3} \) with label \( y \in \{1, \dots, Z\} \), an adversarial attack aims to mislead a 3D DNN classifier \( f \) by perturbing \( \mathcal{P} \) to generate an adversarial point cloud \( \mathcal{P}^{'} \), i.e., \( \mathcal{P}^{'} = \mathcal{P} + \sigma \), where \( \sigma \) represents a carefully crafted perturbation. The adversarial point cloud \( \mathcal{P}^{'} \) is obtained by solving the following optimization problem, commonly via gradient descent:  
\begin{equation}
\min_{\mathcal{P}^{'}} \Bigl( L_{\text{mis}}(f, \mathcal{P}^{'}, y) + \lambda_1 D_{\mathcal{I}}\bigl(\mathcal{P}, \mathcal{P}^{'}\bigr) \Bigr),
\label{eq:basic}
\end{equation}
where \( L_{\text{mis}}(\cdot) \) promotes misclassification (e.g., negative cross-entropy loss), \( D_{\mathcal{I}}(\cdot,\cdot) \) enforces imperceptibility constraints, e.g., Chamfer distance and Hausdorff distance, and \( \lambda_1 \) is a weighting parameter. 
This study primarily focuses on untargeted attacks, and targeted attacks are also possible.

\begin{figure*}[!t]
\centering
\includegraphics[width=1\linewidth]{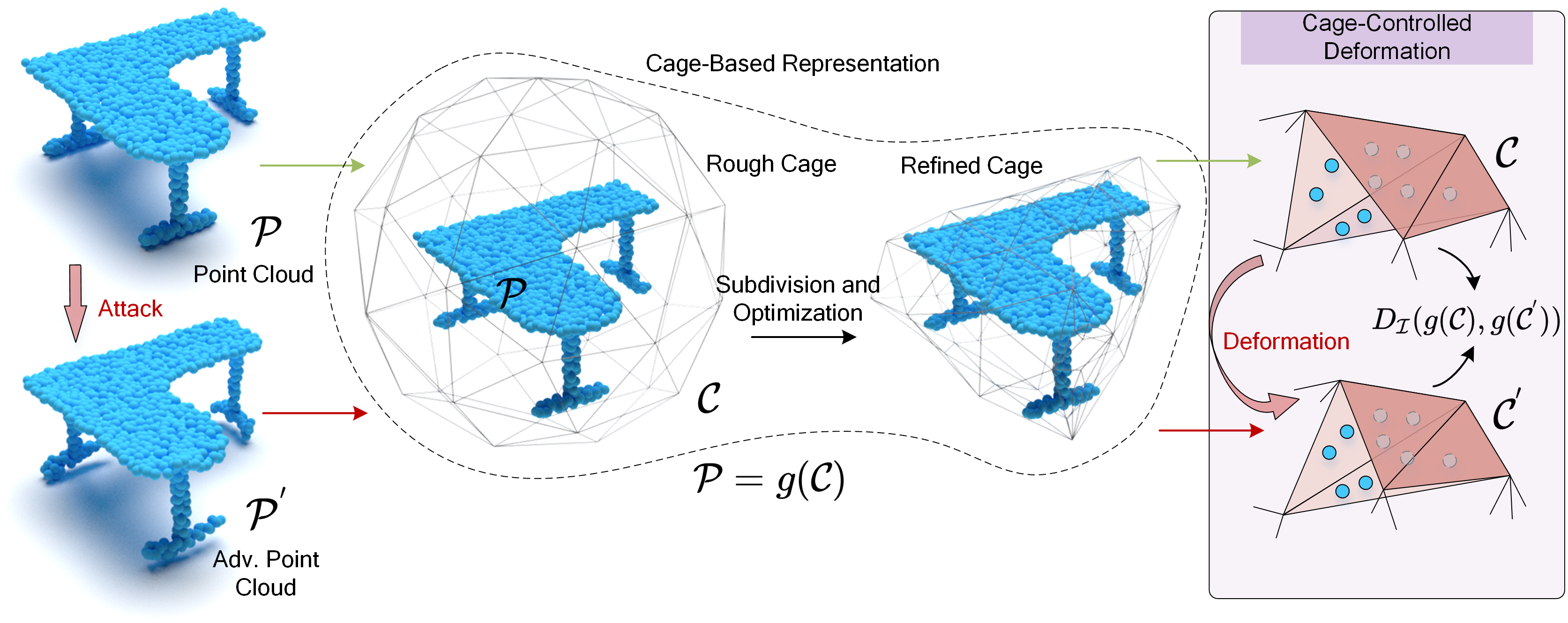} 
\caption{
Illustration of our \OurModel{} framework. 
Given an input point cloud, a surrounding cage is first constructed, followed by subdivision and vertex optimization. By leveraging the structured nature of the cage, controlled shape deformations are blended naturally into the object's inherent geometry, ensuring the plausibility of the adversarial point cloud.
}
\label{fig:framework}
\end{figure*}

\firstpara{Deformation-Based Adversarial Attacks}
Strictly enforcing isolated point-wise constraints often hinders transferability and undefendability, motivating an alternative approach: applying global shape deformations. By introducing modifications \( \operatorname{Deform}(\cdot) \) to the overall shape,  the adversarial point cloud is  obtained as:
\begin{equation}
\mathcal{P}^{'} = \operatorname{Deform} ( \mathcal{P}).
\end{equation}

Deformation can be achieved by stretching the shape's parameterized plane~\cite{Tang-2023-ManifoldAttack} or propagating deformations from selected points~\cite{lou2024hide}. However, these methods primarily deform the shape by manipulating structure-independent control points, often leading to unnatural distortions that compromise the plausibility of the adversarial point cloud.

We argue that a more structured deformation approach could induce changes blending naturally with the object's shape, ensuring adversarial modifications remain plausible.

\subsection{Our Structured Deformation Solution}  

To facilitate structured deformation, we employ a cage~\cite{nieto2012cage,garcia2013cages,stroter2024-CageSurvey} as an intermediary representation that encodes the shape of the point cloud and guides its transformation.

\firstpara{Cage-Based Point Cloud Representation}  
Let \(\mathcal{C} = \{c_j\}_{j=1:m}\) denote the cage that forms a boundary around the target object \(\mathcal{P}\). The point cloud \(\mathcal{P}\) can be expressed in terms of barycentric coordinates relative to the
control points of the cage, e.g.,
cage vertices, using the mapping function $g$:
\begin{equation}
\mathcal{P} = g(\mathcal{C}).
\end{equation}
Specifically, each point \(p_i\) within \(\mathcal{P}\) is represented as:
\begin{equation}
p_i = \sum_{j=1}^{m} \lambda_{i,j} c_j,
\end{equation}
where \(\lambda_{i,j}\) are barycentric coordinates that satisfy \(\sum_{j=1}^{m} \lambda_{i,j} = 1\) and \(\lambda_{i,j} \geq 0\).

\firstpara{Deformation via Cage Perturbation}  
To  deform \(\mathcal{P}\), we introduce perturbations $\{\Delta c_j\}_{j=1:m} $ to the cage vertices, resulting in the perturbed cage:
\begin{equation}    
\mathcal{C}^{'}=\{c_j^{'}\}_{j=1:m}=\{c_j + \Delta c_j\}_{j=1:m}.
\end{equation}
The deformations are then propagated to the points of the object using~\cite{ju2023mean}, yielding  in a new position \( {p}^{'}_i  \) for each point \( p_i \), which can be calculated as:
\begin{equation}
    {p}^{'}_i = \sum_{j=1}^{m} \lambda_{i,j} (c_j^{'}) = \sum_{j=1}^{m} \lambda_{i,j} (c_j + \Delta c_j).
    \label{eq:propagate}
\end{equation}
By leveraging the structured nature of the cage, our solution ensures that shape modifications naturally blend with the object's inherent geometry, making them less noticeable.

\section{Method}

In this section, we first detail the process of constructing the cage and then introduce our deformation-based point cloud attack approach by applying perturbations to the cage.
Please refer to Fig.~\ref{fig:framework} for demonstration.

\subsection{Cage Construction}

We start by initializing a unit sphere \(\mathcal{S} = \{ \mathcal{C}, \mathcal{T} \}\) that encapsulates the point cloud \(\mathcal{P}\), where \(\mathcal{C}\) represents the cage points, and \(\mathcal{T}\) denotes the set of triangles \(\{t_1, t_2, \ldots, t_n\}\) formed by these points. Each triangle \( t_i \) is defined as \( (c_{i_1}, c_{i_2}, c_{i_3}) \). This sphere is then refined through subdivision and vertex optimization to more accurately match the object's overall contour.

\firstpara{Curvature- and Density-Aware Subdivision}  
To achieve a more precise encapsulation of the point cloud, we introduce the sphere center \(o\) as an auxiliary point to construct tetrahedrons that partition the space around the point cloud, denoted as \(\mathcal{E}=\{e_1, \ldots, e_n\}\), where \(e_i = (o, c_{i_1}, c_{i_2}, c_{i_3})\). A well-constructed cage should ensure that the curvature and density within each tetrahedron \(e_i\) are balanced. Therefore, we define a criterion to determine whether \(e_i\) should be subdivided:
\begin{equation}
{S}(e_i) = S_{cur}(e_i) + \lambda_d S_{den}(e_i),
\end{equation}
where \(S_{cur}(e_i)\) and \(S_{den}(e_i)\) represent the average curvature and density of the points within the tetrahedron, and \(\lambda_d\) is a weighting parameter. If \({S}(e_i)\) exceeds a threshold \(\tau\), we subdivide \(e_i\) by splitting the triangle \(t_i\) into four smaller triangles by connecting the midpoints of its edges. 

This subdivision process updates \(\mathcal{C}\) and \(\mathcal{T}\) accordingly, resulting in a more refined cage.

\firstpara{Vertex Optimization}  
The cage obtained from the previous step may not perfectly conform to the point cloud, so we further refine the positions of the cage vertices. Specifically, we optimize the following objective:
\begin{equation}
\small
\mathop{\min}_{\mathcal{C}} \sum_{p_i \in \mathcal{P}} \text{Dist}(p_i, \mathcal{C}) +  \lambda_a \, \text{Var}_{\text{area}}( \mathcal{T}) + \lambda_l \text{Lap} (\mathcal{C}),
\end{equation}
where \(\text{Dist}(p_i, \mathcal{C})\) computes the distance from \(p_i\) to the nearest triangle of the cage, \(\text{Var}_{\text{area}}(\mathcal{T})\) represents the variance in the areas of all triangles, and \(\text{Lap} (\mathcal{C})\) is a regularization term that enforces face smoothing~\cite{zhang20233d}. The parameters \(\lambda_a\) and \(\lambda_l\) are weighting factors that balance each component of the optimization.

Through this process, we obtain a refined cage \(\mathcal{C}\) that adapts to the geometric complexity of the point cloud and closely conforms to its surface, thereby facilitating smooth and natural deformation of the point cloud.
 For simplicity, we continue to denote the refined cage as \(\mathcal{C}\).

\subsection{Attacks via Cage-Based Deformation}  
To achieve a deformation-based adversarial attack that blends naturally into the object’s inherent geometry, instead of directly deforming \(\mathcal{P}\), we introduce perturbations to the cage \(\mathcal{C}\) by solving the following optimization problem:
\begin{equation}
\min_{\mathcal{C}^{'}} L_{\text{mis}} (f, g({\mathcal{C}^{'}}), y) + \lambda_1 D_{\mathcal{I}}(g({\mathcal{C}}), g({\mathcal{C}^{'}})),
\label{eq:final_eq}
\end{equation}
where \(\mathcal{C}^{'}\) represents the perturbed cage. 
We then obtain the final deformed adversarial point cloud \(\mathcal{P}'\) using Eqn.~\ref{eq:propagate}.

The properties of the cage ensure that the applied deformations preserve the natural appearance of the point cloud~\cite{ju2023mean}.  
Therefore, \OurModel{} achieves transferability and undefendability while better balancing plausibility.
\begin{table*}[!t]
\setlength\tabcolsep{1pt}
\centering
\caption{
Comparison on the naturalness of different methods  at their maximum ASR. The evaluation is conducted across different DNN classifiers on  ModelNet40, ScanObjectNN and ShapeNet Part.
}
\label{tab:impercepbility_metrics}
\scalebox{1}{
\begin{tabular}{c|c|cccccc|cccccc|cccccc}
\Xhline{1.0pt} 
\multirow{3}{*}{Model} & \multicolumn{1}{c|}{\multirow{3}{*}{Attack}} & \multicolumn{6}{c|}{ModelNet40} & \multicolumn{6}{c|}{ScanObjectNN} & \multicolumn{6}{c}{ShapeNet Part}\\ \cline{3-20} 
 & \multicolumn{1}{c|}{} & ASR  & CSD & Curv & Uni  & KNN & Lap & ASR   & CSD & Curv & Uni  & KNN & Lap & ASR   & CSD & Curv & Uni & KNN & Lap \\ 
 & \multicolumn{1}{c|}{} & (\%)  & ($10^{-1}$) & ($10^{-3}$) &  & ($10^{-3}$) &  & (\%)  & ($10^{-1}$) & ($10^{-3}$)&  & ($10^{-3}$) &  & (\%)  & ($10^{-1}$) & ($10^{-3}$) &  & ($10^{-3}$) & \\ \hline
\multirow{10}{*}{\rotatebox{90}{PointNet}} 
 & IFGM       & 99.68  & 1.210 & 7.232 & 0.317 & 0.789 & 5.304    & 100.00 & 0.624 & 12.063 & 0.229 & 0.579 & 2.021       & 97.49  & 1.641 & 5.773 & 0.240  & 0.590 & 5.828   \\
 & SI-ADV     & 99.32  & 1.291 & 2.716 & 0.306 & 1.598 & 3.380    & 100.00 & 1.094 & 18.934 & 0.272 & 0.700 & 15.543      & 96.38  & 0.702 & 4.410 & 0.295  & 0.922 & 3.307  \\
 & 3D-ADV     & 100.00 & 1.189 & 5.093 & 0.297 & 0.736 & 2.323    & 100.00 & 0.273 & 6.679 & 0.174 & 0.462 & 1.587        & 100.00  & 0.880 & 4.571 & 0.210  & 0.532 & 3.242 \\
 & KNN-ADV    & 99.68  & 0.775 & 3.947 & 0.369 & 0.677 & 2.122    & 100.00 & 0.861 & 9.442 & 0.171 & 0.303 & 2.010        & 96.40  & 1.509 & 6.949 & 0.272  & \textbf{0.501} & 3.873  \\
 & GeoA$^3$   & 100.00 & 0.439 & 4.368 & 0.292 & 0.717 & 0.961    & 100.00 & 0.444 & 10.172 & 0.207 & 0.502 & 1.525       & 100.00  & 0.681 & 5.908 & 0.266  & 0.689 & 2.054  \\
 & MeshAttack & 96.64  & 2.656 & 5.266 & 0.300 & 0.797 & 7.456    & 100.00 & 17.984 & 14.372 & 0.195 & 0.451 & 38.967     & 97.24  & 1.665 & 3.268 & 0.344  & 0.779 & 13.518  \\
 & $\epsilon$-ISO& 98.66  & 0.433 & 4.815 & 0.305 & 0.745 & 0.934 & 99.72 & 0.297 & 4.662 & 0.188 & 0.417 & \textbf{0.360}& 98.01  & 0.697 & 3.894 & 0.291  & 0.746 & 2.106  \\
 & Mani-ADV   & 93.80  & 16.682 & 18.256 & 0.455 & 1.468 & 35.898 & 98.23 & 14.185 & 82.589 & 0.347 & 1.182 & 34.586      & 96.45  & 16.320 & 9.564 & 0.455  & 1.314 & 40.851  \\
 & HiT-ADV    & 100.00 & 0.986 & 1.282 & 0.288 & 0.745 & 3.475    & 98.86 & 1.022 & 8.129 & 0.168 & 0.424 & 16.854        & 100.00  & 0.688 & 2.115 & \textbf{0.187}  & 0.527 & 2.526  \\
 & Ours       & 100.00 & \textbf{0.429} & \textbf{1.190} & \textbf{0.287} & \textbf{0.667} & \textbf{0.754} & 100.00 & \textbf{0.252} & \textbf{1.251} & \textbf{0.166} & \textbf{0.226} & 1.434 & 100.00  & \textbf{0.625} & \textbf{2.053} & 0.204  & 0.564 & \textbf{2.031}  \\ \hline

\multirow{10}{*}{\rotatebox{90}{PointNet++}}
 & IFGM       & 99.60  & 1.240 & 16.164 & 0.218 & 0.551 & 2.290 &100.00 & 0.413 & 18.380 & 0.147 & 0.364 & 1.019 & 100.00  & 1.432 & 23.517 & 0.272  & 0.734 & 4.514  \\
 & SI-ADV     & 98.87  & 1.761 & 10.816 & 0.256 & 0.677 & 8.746 &100.00 & 1.690 & 3.330 & 0.183 & 0.313 & 9.171 & 98.19 & 1.063 & 10.051 & 0.246  & 0.561 & 6.825  \\
 & 3D-ADV     & 100.00 & 0.768 & 23.120 & 0.205 & 0.614 & 2.903 &100.00 & 0.293 & 21.550 & 0.117 & 0.370 & 1.620 & 100.00  & 1.102 & 19.483 & 0.174  & 0.574 & 4.407  \\
 & KNN-ADV    & 99.92  & 2.435 & 6.106  & 0.169 & 0.578 & 1.877 &100.00 & 1.680 & 17.020 & 0.086 & 0.308 & 2.650 & 99.93  & 2.584 & 9.773 & 0.161  & 0.458 & 2.407  \\
 & GeoA$^3$   & 100.00 & 0.650 & 25.227 & 0.189 & \textbf{0.521} & 3.392 &	100.00 & 0.630 & 3.700 & 0.154 & 0.465 & 1.996 & 100.00 & 0.910 & 19.558 & 0.177  & 0.589 & 4.308  \\
 & MeshAttack & 100.00 & 0.778 & 1.440 & 0.165 & 0.595 & 6.440 &100.00 & 5.060 & 2.430 & 0.113 & 0.290 & 5.290 & 99.86 & 12.447  & 6.184 & 0.240   & 0.918 & 19.021  \\
 & $\epsilon$-ISO & 98.70  & 0.593 & 13.973 & \textbf{0.159} & 0.538 & 1.152 &99.35 & 0.591 & 28.430 & 0.108 & 0.326 & \textbf{1.130} & 98.82  & 0.774 & 14.751 & \textbf{0.152}   & 0.455 & 1.711 \\
 & Mani-ADV   & 94.98  & 16.906 & 18.215 & 0.455 & 1.522 & 27.047 &95.80 & 5.710 & 73.830 & 0.351 & 0.988 & 29.080 & 91.06 & 16.720 & 11.109 & 0.455  & 1.555 & 33.604  \\
 & HiT-ADV    & 100.00 & 0.628 & 2.799 & 0.180 & 0.606 & 5.177 &98.83 & 0.392 & 2.866 & \textbf{0.103} & 0.308 & 12.050& 100.00  & 2.895 & 15.953 & 0.164  & 0.497 & 16.444  \\
 & Ours       & 100.00 & \textbf{0.419} & \textbf{1.191} & 0.185 & 0.594 & \textbf{0.608} &100.00 & \textbf{0.240} &\textbf{1.053} & 0.106 & \textbf{0.288} & 1.219 & 100.00  & \textbf{0.666} & \textbf{4.493} & 0.200  & \textbf{0.421} & \textbf{1.659}  \\ \hline

 \multirow{10}{*}{\rotatebox{90}{DGCNN}} 
 & IFGM       & 98.71  & 2.528 & 27.114 & 0.298 & 0.785 & 5.630      & 99.97 & 0.454 & 17.191 & 0.176 & 0.497 & 2.625     & 99.51  & 2.324 & 35.788 & 0.248 & 0.759 & 9.266  \\
 & SI-ADV     & 96.08  & 2.574 & 12.627 & 0.555 & 1.599 & 11.128     & 98.68 & 2.620 & 53.978 & 0.303 & 0.903 & 14.953    & 96.43  & 1.445 & 19.587 & 0.246 & 0.682 & 12.487 \\
 & 3D-ADV     & 100.00 & 1.713 & 36.542 & 0.292 & 0.868 & 6.136      & 100.00 & 0.460 & 20.537 & 0.164 & 0.485 & 2.585    & 100.00  & 1.306 & 17.646 & 0.189 & 0.684 & 6.138 \\
 & KNN-ADV    & 96.15  & 5.911 & 13.900 & 0.399 & 0.698 & 8.637      & 100.00 & 2.742 & 36.153 & 0.181 & 0.473 & 17.648   & 98.09  & 7.300 & 28.983 & 0.237 & 0.635 & 17.191 \\
 & GeoA$^3$   & 100.00 & 2.014 & 39.848 & 0.289 & 0.865 & 7.745      & 99.77 & 1.870 & 69.196 & 0.287 & 0.913 & 12.244    & 100.00  & 1.160 & 24.029 & 0.189  & 0.718 & 4.261 \\
 & MeshAttack & 100.00 & 1.126 & 1.981  & 0.319 & 0.837 & 2.625      & 100.00 & 8.034 & 14.622 & 0.191 & 0.464 & 21.737   & 99.43   & 1.480 & 7.100 & 0.223  &  0.709 & 17.106 \\
 & $\epsilon$-ISO & 98.13  & 0.667 & 13.002 & 0.300 & 0.838 & 1.940  & 98.30 & 0.855 & 39.491 & 0.193 & 0.574 & 5.404     & 100.00  & 1.077 & 18.702 & 0.217 & 0.710 & \textbf{2.969} \\
 & Mani-ADV   & 97.45  & 16.369 & 22.059 & 0.455 & 1.802 & 44.680    & 99.67 & 14.512 & 81.664 & 0.351 & 1.231 & 16.514   & 95.20  & 16.957 & 16.301 & 0.455  & 1.181 & 46.438 \\
 & HiT-ADV    & 100.00 & 0.595 & 1.565  & 0.287 & 0.837 & 2.183      & 98.80 & 0.770 & 6.539 & 0.166 & 0.441 & 6.960      & 100.00   & 1.075 & 6.920 & \textbf{0.187} & 0.632 & 3.633 \\
 & Ours       & 100.00 & \textbf{0.524} & \textbf{1.482}  & \textbf{0.286} & \textbf{0.697} & \textbf{0.872} & 100.00 & \textbf{0.235} & \textbf{1.277} & \textbf{0.164} & \textbf{0.419} & \textbf{1.923} & 100.00   & \textbf{1.016} & \textbf{6.920} & 0.206  & \textbf{0.571} & 3.185 \\ \hline


 \multirow{10}{*}{\rotatebox{90}{PointMLP}}
 & IFGM       & 99.69 & 0.686 & 10.895 & 0.334 & 0.818 & 2.071 &99.93 & 1.836 & 11.899 & 0.219 & 0.550 & 1.923 & 95.39 & 1.828 & 22.607 & 0.264 & 0.662 & 8.726  \\
 & SI-ADV     & 99.14 & 4.227 & 43.038 & 0.434 & 1.468 & 10.664 &99.80 & 3.897 & 12.632 & 0.304 & 0.708 & 10.861 & 96.98 & 4.833 & 52.695 & 0.374 & 1.169 & 6.554  \\
 & 3D-ADV     & 100.00 & 0.665 & 3.716 & 0.314 & 0.862 & 1.539 &100.00 & 0.872 & 3.572 & \textbf{0.142} & 0.519 & 2.843 & 100.00 & 1.242 & 27.141 & 0.260 & 0.778 & 7.237  \\
 & KNN-ADV    & 94.27 & 1.280 & 2.807 & 0.294 & \textbf{0.752} & 1.730 &100.00 & 0.928 & 7.514 &0.173  & 0.414 & 2.398 & 91.88 & 8.398 & 24.732 & 0.236 & 0.645 & 12.106  \\
 & GeoA$^3$   & 98.75 & 1.461 & 30.870 & 0.381 & 0.995 & 8.522 &99.60 & 1.373 & 4.637 & 0.304 & 0.923 & 2.970 & 96.48 & 1.771 & 18.628 & 0.238 & 0.692 & 6.418  \\
 & MeshAttack & 100.00 & 6.083 & 26.496 & 0.323 & 1.156 & 15.901 &100.00 & 37.670 & 43.701 & 0.211 & 0.576 & 12.550 & 100.00 & 30.037 & 25.680 & 0.383 & 1.371 & 17.336  \\
 & $\epsilon$-ISO & 95.54 & 0.646 & 9.860 & 0.325 & 0.835 & 1.527 &97.85 & 0.656 & 19.460 & 0.212 & 0.522 & \textbf{2.333} & 93.54 & 1.609 & 8.595 & 0.267 & 0.663 & 6.577 \\
 & Mani-ADV   & 95.99 & 15.823 & 48.741 & 0.287 & 0.855 & 15.204 &95.00 & 15.101 & 93.280 & 0.217 & 0.584 & 10.414& 92.79 & 44.199 & 49.923 & 0.235 & 0.856 & 18.628  \\
 & HiT-ADV    & 91.02 & 8.285 & 26.596 & 0.282 & 0.794 & 15.822 &	92.81 & 9.153 & 2.580 & 0.157 & 0.417 & 6.030 & 90.63 & 17.282 & 46.457 & 0.223 & 0.672 & 14.613  \\
 & Ours       & 100.00 & \textbf{0.638} & \textbf{2.409} & \textbf{0.280} & 0.768 & \textbf{1.257} &100.00 & \textbf{0.654} & \textbf{2.176} & 0.165 & \textbf{0.359} & 2.448 & 100.00 & \textbf{1.240} & \textbf{8.535} & \textbf{0.204} & \textbf{0.630} & \textbf{5.996} \\  \hline

 \multirow{10}{*}{\rotatebox{90}{PCT}}
 & IFGM       & 92.94  & 2.238 & 8.015 & 0.272 & 0.488 & 3.242 &93.73 & 0.741 & 9.022 & 0.174 & 0.342 & 2.396 & 96.25 & 1.046 & 7.803 & 0.219 & 0.754 & 2.168  \\
 & SI-ADV     & 100.00  & 5.142 & 26.590 & 0.340 & 0.888 & 40.391 &99.81 & 4.602 & 81.945 & 0.229 & 0.437 & 15.626 & 88.46 & 5.180 & 32.986 & 0.287 & 0.880 & 40.168  \\
 & 3D-ADV     & 100.00 & 0.754 & 7.404 & 0.280 & 0.682 & 2.417 &100.00 & 0.484 & 7.044 & 0.161 & 0.399 & 2.352 & 96.39 & 1.284 & 11.353 & 0.218 & 0.538 & 3.717  \\
 & KNN-ADV    & 98.96  & 5.271 & 13.244 & 0.282 & 0.522 & 16.738 &100.00 & 3.945 & 34.081 & 0.267 & 0.285 & 24.168 & 100.00 & 5.442 & 17.748 & 0.239 & \textbf{0.370} & 18.984  \\
 & GeoA$^3$   & 98.75 & 0.441 & 1.751 & 0.233 & 0.602 & 3.880 &98.82 & 0.423 & 3.415 & 0.192 & 0.548 & 6.214 & 94.38 & 1.315 & 4.671 & 0.192 & 0.686 & 2.729  \\
 & MeshAttack & 97.92 & 2.763 & 16.743 &\textbf{0.228}&\textbf{0.115} & 19.418 & 99.74 & 17.984 & 82.589 & 0.147 & 0.597 & 49.532 & 100.00 & 18.859 & 23.738 & 0.265 & 0.834 & 43.740  \\
 & $\epsilon$-ISO & 93.75 & 1.912 & 8.986 & 0.270 & 0.433 & 3.041 &94.46 & 1.902 & 18.651 & 0.171 & 0.276 & 4.601 & 92.19 & 2.141 & 10.722 & 0.227 & 0.605 & 3.368 \\
 & Mani-ADV   & 94.79 & 56.709 & 40.141 & 0.316 & 0.554 & 39.001 &93.00 & 17.601 & 79.120 & 0.234 & 0.394 & 35.423 & 85.46 & 40.546 & 35.053 & 0.193 & 0.381 & 33.291  \\
 & HiT-ADV    & 98.75 & 15.735 & 31.040 & 0.300 & 0.253 & 36.570 &97.65 & 1.737 & 2.518 & 0.165 & \textbf{0.237} & 28.145 & 97.50 & 18.632 & 43.231 & 0.205 & 0.394 & 44.499  \\
 & Ours       & 100.00  & \textbf{0.344} & \textbf{1.133} & 0.267 & 0.759 & \textbf{1.057} & 100.00 & \textbf{0.382} & \textbf{1.044} & \textbf{0.153} & 0.348 & \textbf{2.144} & 100.00 & \textbf{0.608} & \textbf{3.457} & \textbf{0.188} & 0.572 & \textbf{1.960} \\  \hline
 \multirow{10}{*}{\rotatebox{90}{Mamba3D}}
 & IFGM           & 97.73 & 0.919 & 11.345 & 0.314 & 0.869 & \textbf{1.410}    & 99.13 & 0.665 & 17.802 & 0.187 & 0.543 & 1.334 &98.43  & 0.792  & 14.574  & 0.251 & 0.706 & 1.372          \\
 & SI-ADV         & 100.00 & 1.003 & 10.990 & 0.355 & 0.832 & 12.120  & 99.91 & 0.624 & 12.465 & 0.227 & 0.599 & 9.542          &99.96  & 0.814  & 11.728  & 0.291 & 0.716 & 10.831         \\
 & 3D-ADV         & 95.51 & 0.638 & 9.443 & 0.328 & 0.813 & 1.561     & 94.52 & 0.339 & 13.223 & 0.164 & 0.475 & 1.318          &95.02  & 0.489  & 11.333  & 0.246 & 0.644 & 1.440          \\
 & KNN-ADV        & 100.00 & 1.275 & 5.648 & 0.288 & 0.780 & 1.960    & 99.74 & 0.929 & 8.667 & 0.162 & 0.391 & 1.971           &99.87  & 1.102  & 7.158   & 0.225 & 0.586 & 1.966          \\
 & GeoA$^3$       & 100.00 & 1.312 & 22.814 & 0.356 & 0.898 & 4.829   & 100.00 & 0.831 & 2.047 & 0.212 & 0.727 & 7.543          &100.00 & 1.072  & 12.431  & 0.284 & 0.813 & 6.186          \\
 & MeshAttack     & 99.87 & 15.794 & 21.899 & 0.290 & 0.809 & 8.536 & 100.00 & 17.193 & 15.526 & 0.190 & 0.479 & 19.123         &99.94  & 16.494 & 18.713  & 0.240 & 0.644 & 13.830         \\
 & $\epsilon$-ISO & 99.22 & 0.910 & 12.085 & 0.320 & \textbf{0.739} & 1.682    & 99.81 & 0.527 & 21.328 & 0.182 & 0.528 & 1.887 &99.52  & 0.719  & 16.707  & 0.251 & 0.634 & 1.785           \\
 & Mani-ADV       & 94.48 & 22.292 & 27.450 & 0.511 & 1.989 & 26.836  & 92.55 & 14.050 & 82.126 & 0.324 & 1.048 & 15.268        &93.52  & 18.171 & 54.788  & 0.418 & 1.519 & 21.052          \\
 & HiT-ADV        & 99.85 & 1.515 & 5.754 & 0.299 & 0.801 & 9.842     & 95.78 & 1.166 & 12.323 & \textbf{0.153} & 0.399 & 14.240&97.82  & 1.341  & 9.039   & 0.226 & 0.600 & 12.041         \\
 & Ours           & 100.00 & \textbf{0.566} & \textbf{1.934} & \textbf{0.282} & 0.810 & 2.330   & 100.00 & \textbf{0.277} & \textbf{1.605} & 0.160 & \textbf{0.380} & \textbf{1.231} &100.00 & \textbf{0.422} & \textbf{1.770} & \textbf{0.221} & \textbf{0.595} & \textbf{1.781}\\
 \Xhline{1.0pt}
\end{tabular}
 }
\end{table*}
\section{Experimental Results}

\subsection{Experimental Setup}


\firstpara{Implementation Details}
We implement the \OurModel{} framework using PyTorch~\cite{Pytorch}. For each point cloud, we initialize a unit sphere to envelope it. The cage is then subdivided by calculating the density score and mean curvature for each tetrahedron, normalizing these values to the range [0,1]. We compute the subdivision criterion \(S(\cdot)\) as a weighted sum with \(\lambda_d = 0.25\), and set the subdivision threshold \(\tau\) such that the top one-fifth of the tetrahedrons require subdivision.
During the vertex optimization process, we perform 2000 iterations to determine the optimal positions of the cage vertices. We set \(\lambda_a = 10.0\) to maintain the area of triangles, while \(\lambda_l = 100.0\) serves as a regularization term for surface smoothness. 
We solve the Eqn.~\ref{eq:final_eq}  in  \OurModel{}  following the approach of C\&W~\cite{Carlini-2017-cw}, with \(\lambda_1 = 1.0\). All experiments are conducted on a workstation equipped with dual 2.40 GHz CPUs, 128 GB of RAM, and eight NVIDIA RTX 3090Ti GPUs.

\firstpara{Datasets} 
We evaluate our approach on three publicly available datasets: ModelNet40~\cite{wu20153d}, ShapeNet Part~\cite{chang2015shapenet}, and the real-world ScanObjectNN~\cite{uy-scanobjectnn-iccv19}. 
Following~\cite{Xiang-2019-3DAdversarialPCD}, each point cloud is randomly sampled to 1,024 points. 

\firstpara{Victim 3D DNN Classifiers}  
We evaluate our approach on six representative victim models with diverse architectures, including PointNet~\cite{Qi-2017-Pointnet}, PointNet++\cite{Qi-2017-Pointnet++}, DGCNN\cite{Wang-2019-DGCNN}, PointMLP~\cite{PointMLP}, Point Transformer (PCT)\cite{PT}, and the recent Mamba3D\cite{han2024mamba3d}.
All models are trained following the protocols specified in their original publications.

\firstpara{Baseline Attack Methods}
We select nine baseline attack methods for comparison, including traditional approaches such as the gradient-based method IFGM~\cite{Dong-2020-FGM_PGD_IMPLEMENT}, the direction-based method SI-ADV~\cite{huang2022shape}, and optimization-based methods like 3D-ADV~\cite{Xiang-2019-3DAdversarialPCD} and GeoA$^3$~\cite{wen2020geometry}. Additionally, we consider several deformation-based attack methods, including KNN-ADV~\cite{tsai2020robust}, MeshAttack~\cite{zhang20233d}, and $\epsilon$-ISO~\cite{dong2022isometric}, as well as Mani-ADV~\cite{Tang-2023-ManifoldAttack} and HiT-ADV~\cite{lou2024hide}. This diverse selection of attacks provides a robust baseline for validating the effectiveness of our approach.

\begin{table*}
\setlength\tabcolsep{2.7pt}
\centering
\caption{Transferability performance of different attack methods. The transferability is measured by
  attack success rate (\%)  on target models using adversarial examples that are generated  for attacking source models.}
\label{tab:Transferability}
\scalebox{1.0}{
\begin{tabular}{c|c|c|cccccccccc}
\Xhline{1.0pt}
Data  & Source & Target & IFGM  & SI-ADV & 3D-ADV & KNN-ADV & GeoA$^3$ & MeshAttack & $\epsilon$-ISO & Mani-ADV & HiT-ADV & Ours\\ \hline
\multirow{10}{*}{\rotatebox{90}{ModelNet40}}   & \multirow{5}{*}{PointNet} & DGCNN     & 39.46 & 31.98 & 16.85 & 20.34 & 21.88 & 60.74   & 19.53 & 61.87   & 11.10 & \textbf{68.55}  \\
                               &                           & PointNet++         & 24.29 & 22.34 & 11.68 & 13.85 & 14.47 & \textbf{74.02} & 13.28 & 65.28   & 8.67  & 66.01 \\
                               &                           & PointMLP  & 15.31 & 11.49 & 8.51  & 13.75 & 11.06 & 5.83           & 12.50 & \textbf{59.22} & 8.79  & 20.54      \\
                               &                           & PCT  & \textbf{54.97} & 33.98 & 28.52 & 26.10 & 23.08 & 31.82 & 30.53 & 40.40 & 17.06 & 43.75       \\ 
                               &                           & Mamba3D   & 49.59 & 38.36 & 17.31 & 23.70 & 12.75 & 56.03 & 19.26 & \textbf{67.75} & 12.05 & 55.12 \\ \cline{2-13}
                               & \multirow{5}{*}{DGCNN}    & PointNet  & 25.95 & 32.32 & 24.01 & 19.21 & 42.45 & 75.42   & 27.38 & 48.50   & 11.75 & \textbf{78.00}  \\
                               &                           & PointNet++         & 21.02 & 38.17 & 25.99 & 18.29 & 50.78 & 83.75    & 30.06 & 65.60     & 11.18 & \textbf{84.54} \\
                               &                           & PointMLP  & 35.65 & 21.43 & 16.96 & 14.58 & 13.28 & 25.89          & 26.44 & \textbf{46.08} & 31.27 & 21.25           \\
                               &                           & PCT  & 28.35 & 33.74 & 24.48 & 20.45 & 15.91 & \textbf{48.81} & 27.84 & 48.62 & 18.88 & 47.92          \\ 
                               &                           & Mamba3D   & 41.36 & 41.73 & 44.32 & 44.38 & 19.32 & 78.64 & 46.59 & \textbf{79.96} & 30.42 & 53.13 \\ \hline
                               
\multirow{10}{*}{\rotatebox{90}{ScanObjectNN}} & \multirow{5}{*}{PointNet} & DGCNN     & 43.16 & 50.78 & 27.78 & 38.19 & 36.13 & 67.86 & 30.51 & \textbf{78.59} & 19.71 & 68.44 \\
                               &                           &PointNet++&26.57	& 35.47	& 19.26	& 26.00	& 23.89	& 82.70	& 20.75	& \textbf{82.92}	& 15.40	& 65.90\\
                               &                           &PointMLP&16.75	& 18.24	& 14.03	& 25.82	& 18.26	& 26.51	& 19.53	& \textbf{75.22}	& 15.61	& 20.51\\
                               &                           &PCT&30.12	& 35.55 & 47.02	& 49.00	& 38.11	& 51.96	& 47.69	& 51.32	& 30.29	& \textbf{53.68}\\
                               &                           & Mamba3D   & 58.30 & 63.64 & 41.10 & 52.08 & 53.34 & 45.71 & 44.33 & 73.90 & 28.37 & \textbf{77.15} \\ \cline{2-13}
                               & \multirow{5}{*}{DGCNN}    & PointNet  & 39.86 & 66.71 & 30.47 & 49.31 & 57.99 & 69.57 & 54.89 & 69.22 & 16.52 & \textbf{75.18} \\
                               &                           &PointNet++&32.29	& 78.78	& 32.98	& 46.95	& 69.37	& 77.25	& 60.26	& \textbf{93.63}	& 15.72	& 81.48\\
                               &                           &PointMLP&54.76	& 44.23	& 21.52	& 37.43	& 18.14	& 43.96	& 53.01	& \textbf{65.77}	& 23.88	& 37.48\\
                               &                           &PCT&43.55	& 49.64	& 31.07	& 22.49	& 21.73	& \textbf{69.39}	& 55.81	& 65.02	& 26.54	& 46.19\\
                               &                           & Mamba3D   & 56.74 & 58.64 & 45.05 & 60.42 & 65.33 & 58.15 & 71.20 & 73.14 & 12.50 & \textbf{76.96} \\  \hline
\multirow{10}{*}{\rotatebox{90}{Shapenet Part}} & \multirow{5}{*}{PointNet} & DGCNN     & 28.82 & 5.38  & 2.92  & 3.37  & 3.98  & \textbf{32.14} & 1.63  & 23.94   & 14.90 & 24.13  \\
                               &                           &  PointNet++        & 29.17 & 9.64  & 4.31  & 6.41  & 7.01  & \textbf{51.79} & 4.69  & 23.45          & 14.93 & 25.17 \\
                               &                           & PointMLP  & 14.87 & 8.86  & 4.86  & 3.13  & 5.00  & 9.72           & 5.36  & \textbf{22.35} & 2.71  & 15.14            \\
                               &                           & PCT  & 14.19 & 8.89 & 13.97 & 11.98 & 9.82 & 12.50 & 17.31 & 14.06 & 2.16 & \textbf{18.30}          \\
                               &                           & Mamba3D  &  34.57 & 	20.71	& 12.67 & 	13.84	& 12.02	& 33.24	& 14.47	& 32.49	& 9.64	& \textbf{35.54}\\ \cline{2-13}
                               & \multirow{5}{*}{DGCNN}    & PointNet  & 51.82 & 20.13 & 40.36 & 17.61 & 50.00 & 56.25          & 49.73 & 21.02          & 18.93 & \textbf{58.55}  \\
                               &                           &  PointNet++        & 47.47 & 25.66 & 29.69 & 8.81  & 32.75 & \textbf{78.13} & 27.99 & 38.62          & 1.15  & 48.03   \\
                               &                           & PointMLP  & 36.10 & 24.74 & 15.87 & 26.04 & 5.56  & 20.58          & 16.15 & \textbf{39.76} & 25.02 & 28.33           \\
                               &                           & PCT  & 25.23 & 12.23 & 22.12 & 12.50 & 5.68 & 30.16 & 26.14 & 15.07 & 4.38 & \textbf{31.25}             \\
                               &                           & Mamba3D   & 38.51	& 22.84	& 26.15	& 29.82	& 20.65	& \textbf{50.47}	& 34.48	& 47.04	& 14.98	&  46.31   \\ 
\Xhline{1.0pt}
\end{tabular}}
\end{table*}

\begin{figure*}[t]
\centering
\includegraphics[width=1\textwidth]{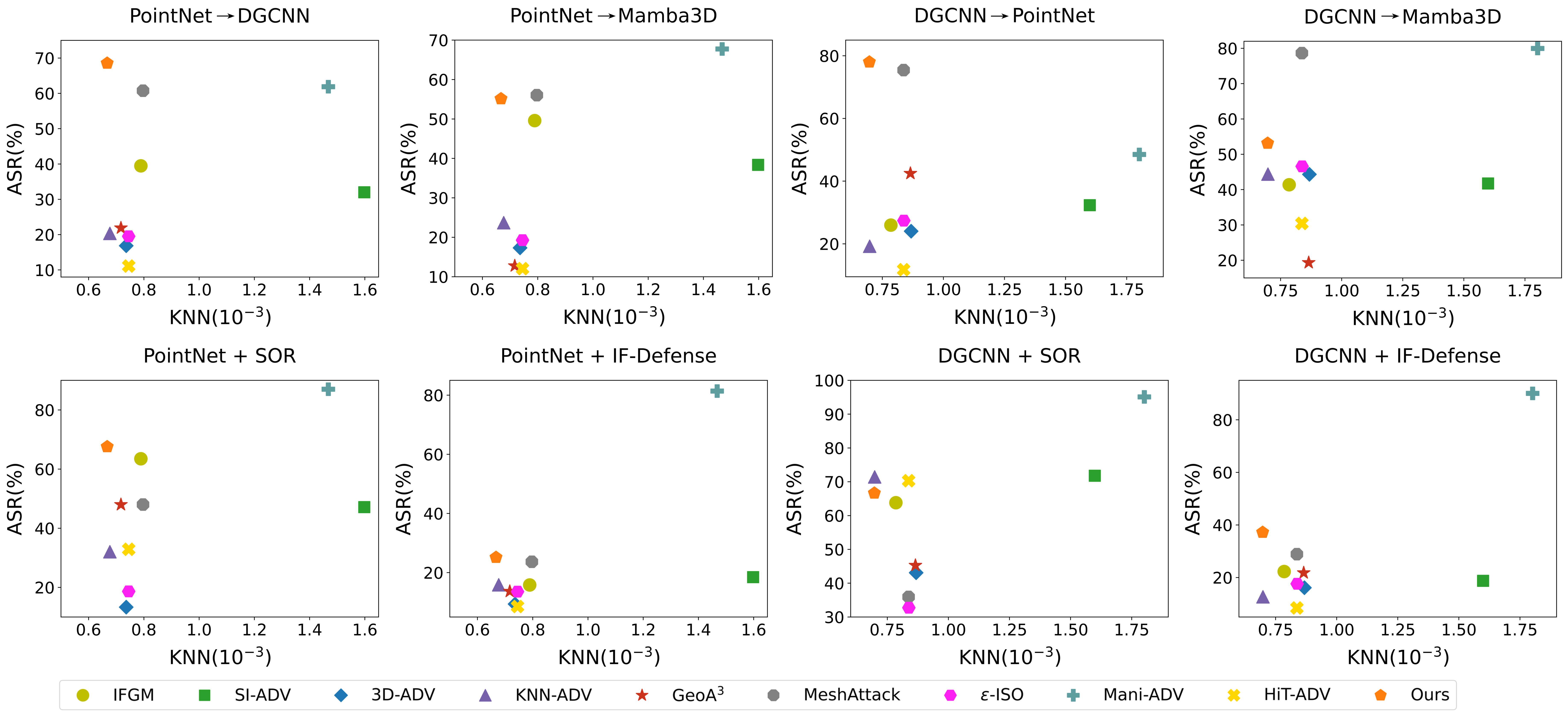}
\caption{
{\textbf{Top row}}: Trade-off between transferability (measured by ASR) and naturalness (measured by KNN) for various attack methods on ModelNet40.
{\textbf{Bottom row}}: Trade-off between undefendability (measured by ASR) and naturalness (measured by KNN) for different attack methods on ModelNet40.
}
\label{fig:tradeoff}
\end{figure*}

\firstpara{Defense Methods}
We adopt four adversarial defense strategies: simple random sampling (SRS), statistical outlier removal (SOR), denoiser and upsampler network (DUP-Net)~\cite{Zhou-2019-dup}, and IF-Defense~\cite{Wu-2020-IFDefense}. The SRS method mitigates attacks by randomly removing 500 points from the input point clouds, while SOR eliminates irregular points that fall outside the mean and standard deviation of the nearest neighbor distances. Building on SOR, DUP-Net enhances point cloud resolution by remapping adversarial samples to the natural manifold. Meanwhile, IF-Defense employs implicit functions to model clean shapes, thereby restoring the integrity of adversarial point clouds.

\firstpara{Evaluation Setting and Metrics}
To ensure fair comparisons, we configure each attack method to achieve its maximum reachable attack success rate (ASR), defined as the proportion of adversarial point clouds that successfully mislead the victim model. Under this condition of maximal adversarial effectiveness, we evaluate the naturalness of the attacks using five widely recognized metrics: curvature standard deviation (CSD)~\cite{lou2024hide}, curvature (Curv), uniform loss (Uni)~\cite{li2019pu}, KNN distance (KNN)~\cite{tsai2020robust}, and Laplacian loss (Lap)~\cite{zhang20233d}. Additionally, we employ ASR  under the same setting for assessing transferability and undefendability.
 Unless explicitly mentioned, all discussions regarding the attack results refer to non-targeted attacks.

\begin{figure*}[!t]
\centering
\includegraphics[width=1\textwidth]{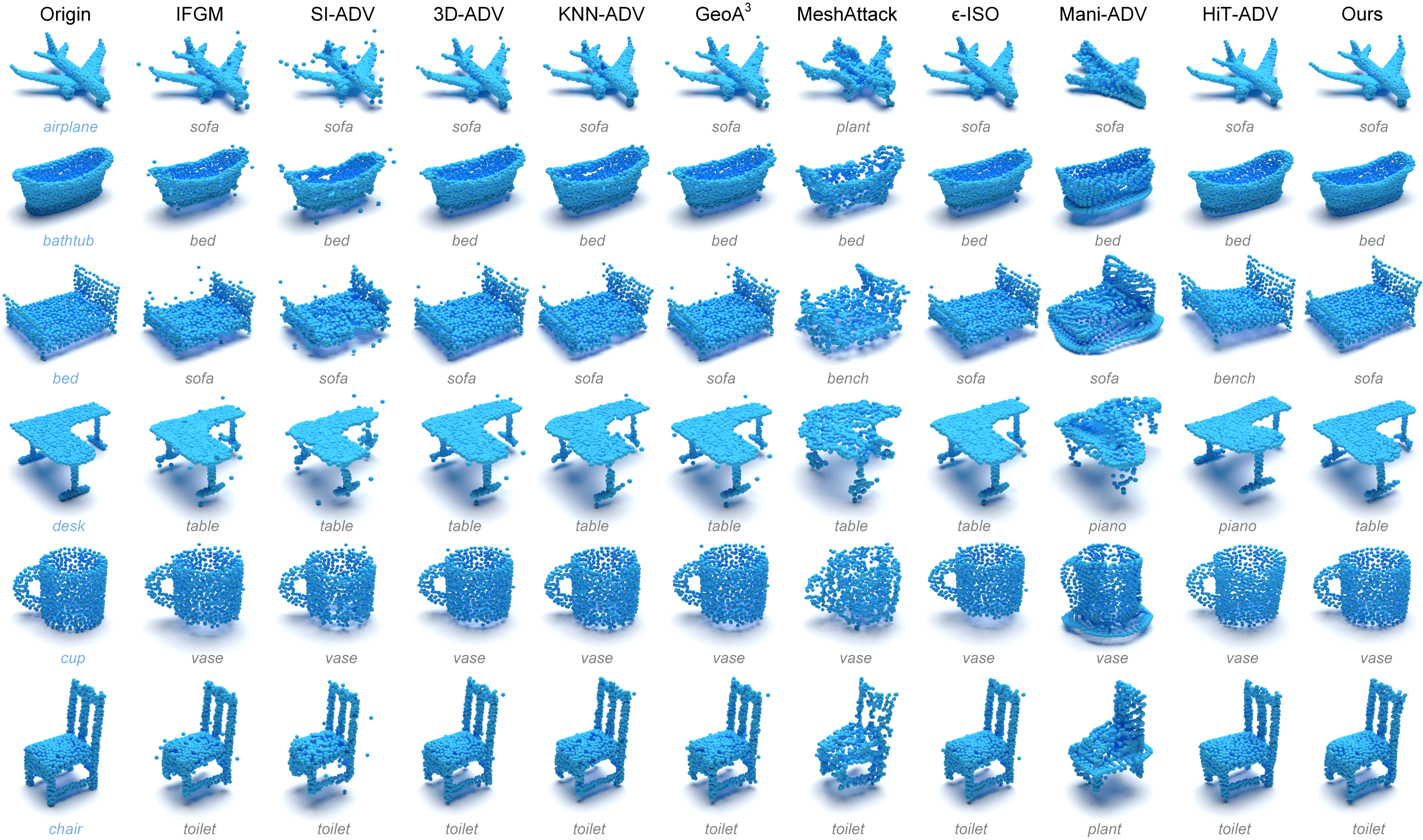}
\caption{Visualizations of original and adversarial point clouds generated to fool PointNet on ModelNet40 by various adversarial attack methods. 
The ground truth and predicted labels are marked in blue and gray below the images.
}
\label{fig:VIS_AdvPC}
\end{figure*}

\begin{table*}
\centering
\caption{Attack success rate (\%)  of different attack methods with and without defense on ModelNet40. The best values are in \textbf{bold}, and the second-best values are \underline{underlined}.}
\vspace{-2mm}
\label{tab:Indefensibility}
\setlength{\tabcolsep}{3.5pt}
\scalebox{1.0}{
\begin{tabular}{c|c|cccccccccc}
\Xhline{1.0pt} 
 Model & Defense & IFGM & SI-ADV & 3D-ADV  & KNN-ADV & GeoA$^3$ & MeshAttack & $\epsilon$-ISO & Mani-ADV & HiT-ADV & Ours\\ \hline
    \multirow{5}{*}{{PointNet}} 
    & -    & 99.68 & 99.32 & \textbf{100.00} & 99.68 & \textbf{100.00} & 96.64 & 98.66 & 93.80 & \textbf{100.00} & \textbf{100.00} \\
    & SRS           & 81.81 & 58.32 & 55.42  & 72.42 & 73.17  & 83.67 & 65.80 & \textbf{93.48} & 31.90 & \underline{85.17} \\
    & SOR           & 63.49 & 47.13 & 13.33  & 32.03 & 47.98  & 48.01 & 18.64 & \textbf{87.00} & 32.89 & \underline{67.58} \\
    & DUP-Net       & 21.49 & 33.12 & 12.92  & 11.32 & 56.73  & \underline{58.46} & 16.49 & \textbf{86.22} & 34.86 & 52.03 \\
    & IF-Defense    & 15.81 & 18.47 & 9.38   & 15.86 & 13.65  & 23.66 & 13.57 & \textbf{81.36} & 8.55  & \underline{25.16} \\\hline
    
    \multirow{5}{*}{{PointNet++}} 
    & -    & 99.60 & 98.87 & \textbf{100.00} & 99.92 & \textbf{100.00} & \textbf{100.00}& 98.70 & 94.98  & \textbf{100.00} & \textbf{100.00} \\
    & SRS           & 63.97 & 72.36 & 54.16  &\underline{ 83.88} & 75.96 & 75.31 & 52.83 & \textbf{94.06} & 32.50  & 71.16 \\
    & SOR           & 69.77 & 75.26 & 23.21  & 72.57 & 47.51 & 48.75 & 34.96 & \textbf{90.71} & 19.58  & \underline{76.61} \\
    & DUP-Net       & 35.00 & 48.95 & 21.72  & 26.17 & 39.30 & \underline{51.25} & 23.99 & \textbf{87.32} & 15.41  & 41.74 \\
    & IF-Defense    & 17.18 & 15.72 & 13.69  & 12.89 & 17.23 & 22.18 & 12.80 & \textbf{83.51} & 3.34   & \underline{23.22} \\ \hline
    
    \multirow{5}{*}{{DGCNN}} 
    & -    & 98.71 & 96.08 & \textbf{100.00}& 96.15 & \textbf{100.00} & \textbf{100.00} & 98.13 & 97.45 & \textbf{100.00}& \textbf{100.00} \\
    & SRS           & 58.62 & 72.17 & 37.85 & \underline{81.25} & 79.65  & 71.87  & 71.71 & \textbf{94.73} & 66.82 & 68.43 \\
    & SOR           & 63.79 & \underline{71.78} & 43.08 & 71.42 & 45.26  & 35.94  & 32.73 & \textbf{95.10} & 70.31 & 66.64 \\
    & DUP-Net       & 56.89 & 69.37 & 38.54 & 74.10 & 53.81  & \underline{85.94}  & 51.78 & \textbf{94.65} & 85.54 & 72.81 \\
    & IF-Defense    & 22.29 & 18.75 & 16.12 & 12.67 & 21.80  & 28.88  & 17.62 & \textbf{90.03} & 8.43  & \underline{37.21} \\ \hline

    
        \multirow{5}{*}{{PointMLP}} 
    & -    & 99.69 & 99.14 & \textbf{100.00} & 94.27 & 98.75 & \textbf{100.00} & 95.54 & 95.99 & 91.02 & \textbf{100.00} \\
    & SRS           & 70.15 & 43.17 & 22.32 & 14.58 & 21.53 & 77.08 & 46.88 & \textbf{90.72} & 37.92 & \underline{77.64} \\
    & SOR           & 70.96 & 44.96 & 25.89 & 26.04 & 13.19 & \underline{87.92} & 23.44 & \textbf{88.75} & 29.17 & 63.89 \\
    & DUP-Net       & 56.62 & 51.44 & 33.93 & 25.00 & 20.14 & \textbf{93.75} & 40.63 & \underline{87.95} & 51.25 & 59.72 \\
    & IF-Defense    & 22.06 & 10.79 & 10.71 & 9.38 & 9.72 & 25.00 & 12.50 & \textbf{84.66} & 3.75 & \underline{31.94} \\  \hline
        \multirow{5}{*}{PCT}
    & -             & 92.94 & \textbf{100.00} & \textbf{100.00} & 98.96 & 98.75 & 97.92 & 93.75 & 94.79 & 98.75 & \textbf{100.00}\\  
    & SRS           & 62.88 & 64.31 & 32.86 & 48.44 & 26.14 & 82.14 & 38.64 & \underline{90.18} & 31.62 & \textbf{91.48} \\ 
    & SOR           & 53.75 & 45.23 & 20.89 & 62.50 & 19.51 & \textbf{95.66} & 24.24 & 90.28 & 35.66 & \underline{92.05} \\ 
    & DUP-Net       & 40.00 & 32.86 & 23.39 & 36.98 & 24.43 & 67.32 & 27.27 & \textbf{89.06} & 43.75 & \underline{85.80} \\ 
    & IF-Defense    & 23.50 & 26.15 & 21.25 & 22.40 & 19.89 & 68.39 & 22.16 & \textbf{86.88} & 13.24 & \underline{71.59} \\ \hline
    \multirow{5}{*}{{Mamba3D}} 
    & -             & 97.73 & \textbf{100.00} & 95.51 & \textbf{100.00} & \textbf{100.00} & 99.87 & 99.22 & 94.48 & 99.85 & \textbf{100.00} \\
    & SRS           & 56.01 & 70.02 & 44.20 & 37.91 & 71.94 & \underline{76.97} & 56.09 & 53.68 & 28.17 & \textbf{77.02} \\
    & SOR           & 40.02 & 50.61 & 17.65 & 68.91 & 38.03 & \textbf{77.73} & 28.21 & 53.88 & 20.96 & \underline{76.92} \\
    & DUP-Net       & 55.44 & 54.31 & 48.09 & 59.46 & 54.04 & \textbf{72.42} & 55.80 & 62.58 & 47.10 & \underline{65.58} \\
    & IF-Defense    & 18.91 & 26.44 & 14.49 & 17.41 & 23.94 & \underline{61.82} & 16.84 & 54.33 & 19.44 & \textbf{63.17} \\
    \Xhline{1.0pt} 
\end{tabular}}
\end{table*}

\subsection{Comparison and Performance Analysis}

\firstpara{Performance on Naturalness}
As shown in Tab.~\ref{tab:impercepbility_metrics}, most attack methods attain high attack success rates, often exceeding 90\% across all six victim classifiers. However, these successful attacks typically come at the cost of perceptual quality—many perturbation-based methods introduce noticeable artifacts, resulting in point clouds that are visually unnatural or structurally distorted. In contrast, deformation-based approaches tend to better preserve the geometric plausibility of the original shapes. Notably, $\epsilon$-ISO, HiT-ADV, and our \OurModel{} consistently outperform other methods in terms of naturalness metrics. Among them, \OurModel{} achieves the best overall scores, indicating that its cage-based deformation strategy is particularly effective in producing adversarial point clouds that remain visually and structurally realistic while maintaining strong attack capability.

\firstpara{Performance on Transferability}
As reported in Tab.\ref{tab:Transferability}, most attack methods exhibit poor transferability: adversarial point clouds crafted on a source model often fail to generalize to unseen target models, with attack success rates commonly dropping below 20\%. In contrast, MeshAttack, Mani-ADV, and our \OurModel{} show consistently higher transferability across different model pairs, achieving success rates up to 70\% in some scenarios.
Among them, MeshAttack performs best in terms of transferability on the ShapeNet Part dataset. However, this comes at the cost of significantly degraded naturalness, producing visibly distorted point clouds. By comparison, \OurModel{} maintains competitive transferability while preserving much higher perceptual quality. The top row of Fig.\ref{fig:tradeoff} visualizes this trade-off, clearly demonstrating that \OurModel{} achieves a more favorable balance between transferability and naturalness than prior approaches.

\firstpara{Performance on Undefendability}
Tab.\ref{tab:Indefensibility} shows that even the simplest defense strategy, SRS, can noticeably reduce the effectiveness of most attacks, typically halving the attack success rate. As stronger defenses are introduced—such as SOR and DUP-Net—the success rates of nearly all high-performing attacks drop substantially. In particular, several deformation-based methods, including $\epsilon$-ISO and HiT-ADV, fall to around 20\% under these defenses in some settings. When applying the strongest defense, IF-Defense, most attacks become almost entirely ineffective. Under this setting, only Mani-ADV and our \OurModel{} consistently achieve attack success rates above 25\% across all three victim models, highlighting their relative robustness. Although \OurModel{} exhibits slightly lower undefendability than Mani-ADV, it significantly outperforms it in terms of perceptual quality and structural plausibility. As illustrated in the bottom row of Fig.\ref{fig:tradeoff}, \OurModel{} offers a more favorable balance between attack effectiveness and naturalness, maintaining visual realism without overly sacrificing robustness.


\begin{figure*}[t]
\centering
\includegraphics[width=1\textwidth]{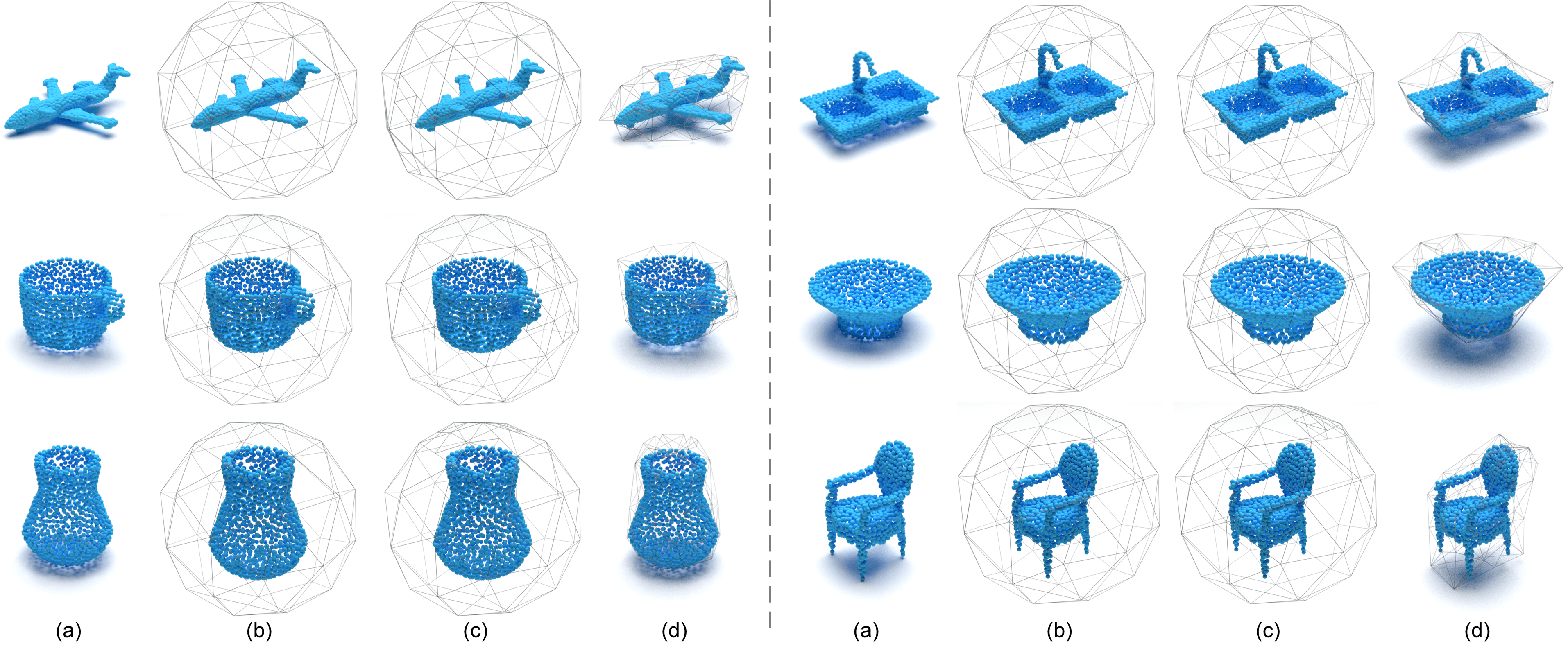}
\caption{Visualizations of (a) original point clouds, (b) initially constructed cages, (c) cages after curvature- and density-aware subdivision, and (d) cages after vertex optimization.}
\label{fig:LatticeVIS}
\vspace{-3mm}
\end{figure*}

\firstpara{Visualization}
We visualize adversarial point clouds generated by various attack methods aimed at fooling PointNet, as shown in Fig.~\ref{fig:VIS_AdvPC}. 
The visualizations reveal that adversarial point clouds produced by most baseline methods exhibit noticeable outliers. In contrast, deformation-based attack methods introduce fewer outliers, though the deformations tend to be more significant. Mani-ADV produces highly conspicuous deformations. HiT-ADV, meanwhile, introduces localized perturbations across select regions, preserving the overall shape yet creating visible distortions in finer details. In comparison, our \OurModel{} achieves the most natural deformations, resulting in the highest level of imperceptibility, making it significantly harder to detect visually.

\begin{figure}[!t]
\centering
\includegraphics[width=1\linewidth]{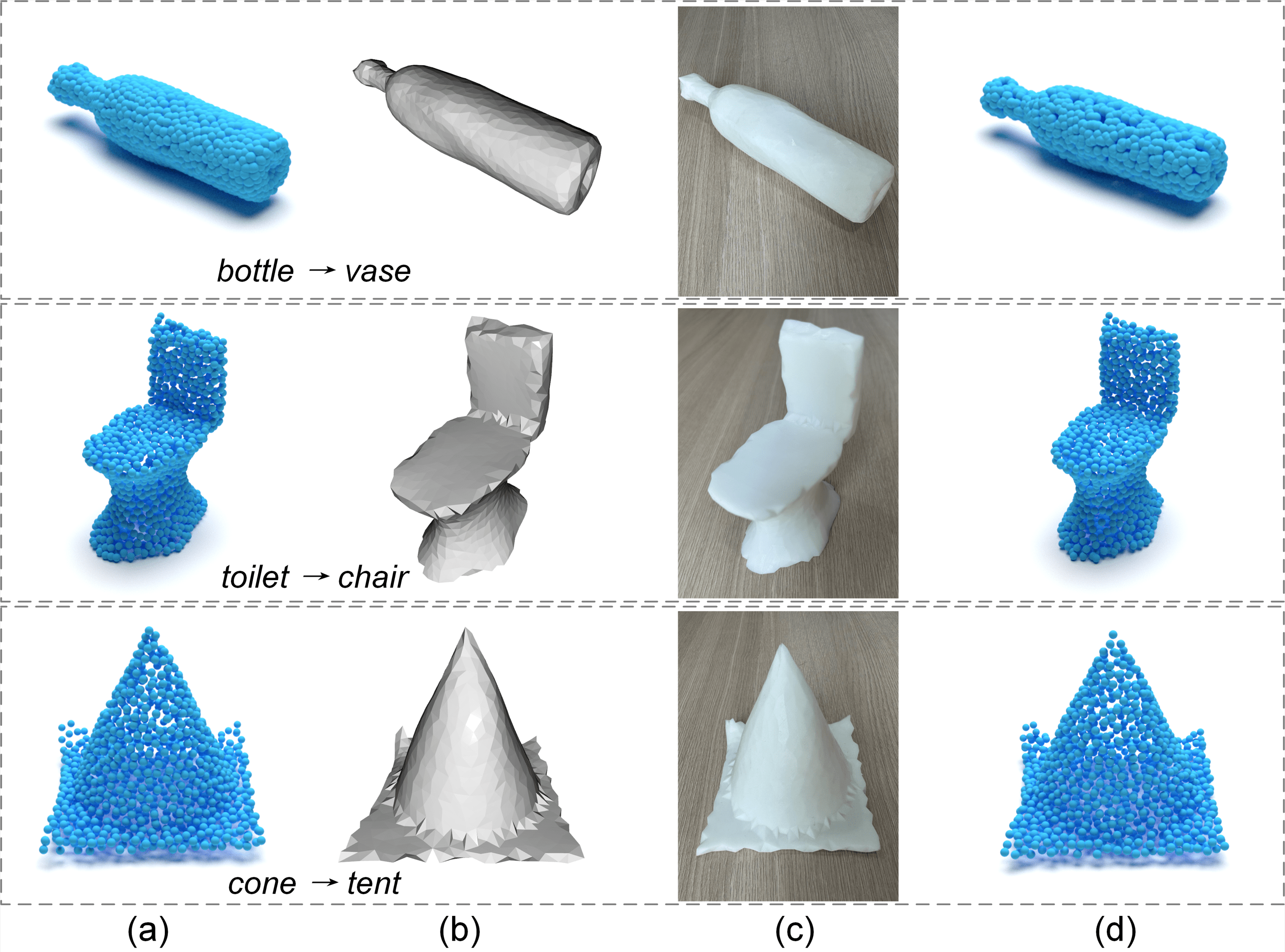} 
\caption{Physical attack targeting PointNet on ModelNet40: (a) Generated adversarial point clouds \(\rightarrow\) (b) reconstructed adversarial meshes \(\rightarrow\) (c) 3D-printed adversarial objects \(\rightarrow\) (d) re-scanned and re-sampled adversarial point clouds.}
\label{fig:Physical}
\end{figure}

\subsection{Ablation Studies and Other Analysis}

\firstpara{Cage Subdivision and Vertex Optimization}
To validate the importance of cage subdivision and vertex optimization in our adversarial attack framework, we visualize the results of point cloud partitioning before and after these operations. As shown in Fig.~\ref{fig:LatticeVIS}, cage subdivision refines larger regions into smaller, more detailed partitions, particularly in complex areas such as object junctions. Vertex optimization further adjusts the vertex positions, ensuring better alignment with the underlying structure of the point cloud and resulting in smoother, more natural deformations. The results in Tab.~\ref{tab:ablation} show that when these operations are omitted, the naturalness metrics degrade, and plausibility decreases. This analysis highlights the critical role that both cage subdivision and vertex optimization play in improving the naturalness and plausibility of our  approach.

\begin{table}[!t]
\centering
\vspace{-1.4mm}
\setlength{\tabcolsep}{1mm}{
\caption{Comparison on the naturalness of different variants of our \OurModel{}, with and without cage subdivision (S) and vertex optimization (O), when attacking PointNet on ModelNet40.
}
\label{tab:ablation}
\scalebox{1}{
\begin{tabular}{cc|cccccc}
\Xhline{1.0pt} 
\multirow{1}{*}{S} & \multirow{1}{*}{O} & ASR  & CSD & Curv & Uni & KNN & Lap  \\  \hline
           &             & 96.59  & 3.183 & 7.797 & 0.296 & 0.810 & 3.389 \\
\checkmark &             & 97.65  & 1.977 & 5.703 & 0.288 & 0.806 & 3.304 \\
           & \checkmark  & \textbf{100.00} & 0.756 & 1.641 & \textbf{0.286} & 0.809 & 1.512 \\
\checkmark & \checkmark  & \textbf{100.00} & \textbf{0.429} & \textbf{1.190} & 0.287 & \textbf{0.667} & \textbf{0.754} \\
\Xhline{1.0pt}
\end{tabular}}
}
\end{table}

\begin{figure*}[!t]
\centering
\includegraphics[width=1.0\textwidth]{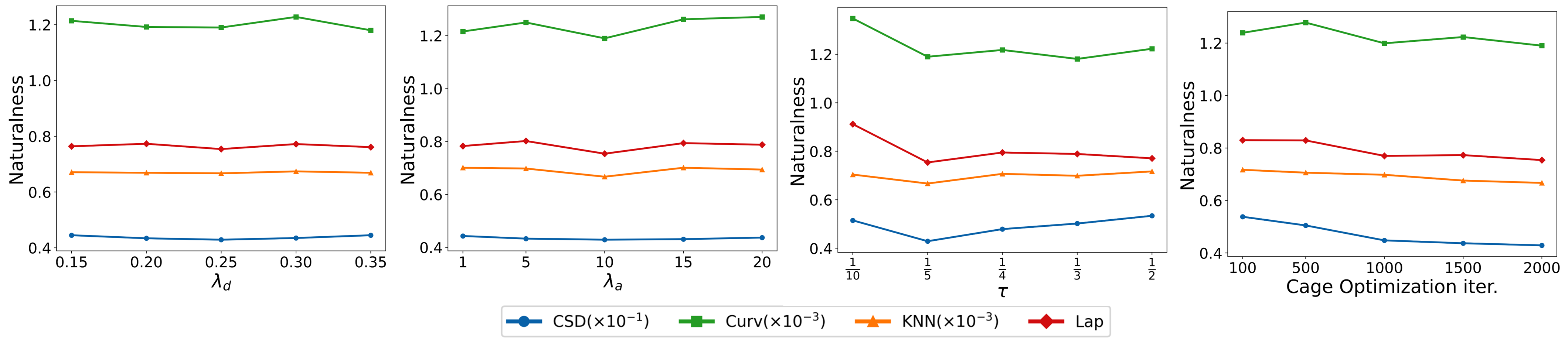}
\caption{Evaluation of the naturalness performance of CageAttack in attacking PointNet trained on the ModelNet40 dataset.
 The performance is measured using the CSD, Curv, KNN, and Lap metrics under different parameter configurations, including the balancing hyperparameters (\(\lambda_d\) and \(\lambda_a\)), the subdivision threshold (\(\tau\)), and the number of iterations for cage optimization.}
\label{fig:TUNING}
\end{figure*}

\firstpara{Physical Attacks} We further validate our approach in a physical attack setting. Specifically, we reconstruct the adversarial point clouds as meshes, 3D-print them into adversarial objects, and then re-scan and re-sample the printed objects to create new input point clouds, which are subsequently fed into the victim model to assess their success in misleading it. Results show that some samples successfully fool the model (see Fig.~\ref{fig:Physical}), demonstrating the potential of our method for physical adversarial attacks.

\firstpara{Parameter Tuning Results}
We analyze the impact of various parameters on the performance of \OurModel{}. The experiments are conducted using PointNet trained on the ModelNet40 dataset, as depicted in Fig.~\ref{fig:TUNING}.  
For the weighting parameter \(\lambda_d\), a value of 0.25 achieves the best balance between \(S_{cur}(\cdot)\) and \(S_{den}(\cdot)\), optimizing the cage subdivision process. Similarly, for \(\lambda_a\), a value of 10.0 provides optimal results for balancing \(\text{Dist}(\cdot,\cdot)\) and \(\text{Var}_{\text{area}}(\cdot)\), ensuring smooth deformations and consistent triangle areas during vertex optimization.
Regarding the subdivision threshold \(\tau\), we evaluate various threshold values and find that subdividing such that the top one-fifth of tetrahedrons require subdivision yields the best results. Further subdivision into finer partitions has a marginal or even adverse effect on performance metrics. For vertex optimization, convergence is typically achieved after 2000 iterations. Therefore, we set the number of iterations to 2000 in our experiments.

\firstpara{User Study on Plausibility}
To evaluate the plausibility of our adversarial attack, we conduct a user study comparing five methods across 100 samples. Ten participants choose, for each sample, the adversarial result that appears most plausible. The results in Tab.~\ref{tab:user} show that adversarial point clouds generated by CageAttack are consistently perceived as more plausible.



\begin{table}[h]
\centering
\vspace{-1mm}
\caption{User study evaluating the plausibility of adversarial point clouds generated by five attack methods. 
}
\label{tab:user}
\scalebox{1}{
\begin{tabular}{ccccc}
\Xhline{1.0pt} 
IFGM        & $\epsilon$-ISO & Mani-ADV & HiT-ADV  &  Ours\\ \hline
1.5\%  & 4.5\%  & 9.3\% &  20.7\% & \textbf{64.0\%}  \\ \Xhline{1.0pt} 
\end{tabular}
}
\vspace{-1mm}
\end{table}
\section{Conclusion}

This paper has proposed a novel cage-based deformation framework for generating adversarial attacks on point clouds. By leveraging the structured nature of cages, our approach enables controlled shape deformations that preserve the naturalness of point clouds. Extensive experiments demonstrate that our method, \OurModel{}, achieves a superior balance between transferability,  undefendability, and plausibility,  compared to state-of-the-art techniques. 
In future work, we aim to further enhance the effectiveness of our approach in the physical domain, e.g., by integrating differentiable simulation and real-world sensor feedback.


{
\bibliographystyle{IEEEtran}
\bibliography{ref}

\begin{thebibliography}{10}
\providecommand{\url}[1]{#1}
\csname url@samestyle\endcsname
\providecommand{\newblock}{\relax}
\providecommand{\bibinfo}[2]{#2}
\providecommand{\BIBentrySTDinterwordspacing}{\spaceskip=0pt\relax}
\providecommand{\BIBentryALTinterwordstretchfactor}{4}
\providecommand{\BIBentryALTinterwordspacing}{\spaceskip=\fontdimen2\font plus
\BIBentryALTinterwordstretchfactor\fontdimen3\font minus \fontdimen4\font\relax}
\providecommand{\BIBforeignlanguage}[2]{{%
\expandafter\ifx\csname l@#1\endcsname\relax
\typeout{** WARNING: IEEEtran.bst: No hyphenation pattern has been}%
\typeout{** loaded for the language `#1'. Using the pattern for}%
\typeout{** the default language instead.}%
\else
\language=\csname l@#1\endcsname
\fi
#2}}
\providecommand{\BIBdecl}{\relax}
\BIBdecl

\bibitem{guo2020deep}
Y.~Guo, H.~Wang, Q.~Hu, H.~Liu, L.~Liu, and M.~Bennamoun, ``Deep learning for 3d point clouds: A survey,'' \emph{TPAMI}, vol.~43, no.~12, pp. 4338--4364, 2020.

\bibitem{han2024mamba3d}
X.~Han, Y.~Tang, Z.~Wang, and X.~Li, ``Mamba3d: Enhancing local features for 3d point cloud analysis via state space model,'' in \emph{ACM MM}, 2024, pp. 4995--5004.

\bibitem{Xiang-2019-3DAdversarialPCD}
C.~Xiang, C.~R. Qi, and B.~Li, ``Generating 3d adversarial point clouds,'' in \emph{CVPR}, 2019, pp. 9136--9144.

\bibitem{Liu-2019-extendingAdv3D}
D.~Liu, R.~Yu, and H.~Su, ``Extending adversarial attacks and defenses to deep 3d point cloud classifiers,'' in \emph{ICIP}, 2019, pp. 2279--2283.

\bibitem{ijcai2021p591}
T.~Bai, J.~Luo, J.~Zhao, B.~Wen, and Q.~Wang, ``Recent advances in adversarial training for adversarial robustness,'' in \emph{IJCAI}, 2021, pp. 4312--4321.

\bibitem{liu2022imperceptible}
D.~Liu and W.~Hu, ``Imperceptible transfer attack and defense on 3d point cloud classification,'' \emph{TPAMI}, vol.~45, no.~4, pp. 4727--4746, 2023.

\bibitem{Hamdi-2020-AdvPC}
A.~Hamdi, S.~Rojas, A.~Thabet, and B.~Ghanem, ``Advpc: Transferable adversarial perturbations on 3d point clouds,'' in \emph{ECCV}, 2020, pp. 241--257.

\bibitem{he2023-PF}
B.~He, J.~Liu, Y.~Li, S.~Liang, J.~Li, X.~Jia, and X.~Cao, ``Generating transferable 3d adversarial point cloud via random perturbation factorization,'' in \emph{AAAI}, vol.~37, no.~1, 2023, pp. 764--772.

\bibitem{chen2024anf}
H.~Chen, S.~Zhao, X.~Yang, H.~Yan, Y.~He, H.~Xue, F.~Qian, and H.~Su, ``Anf: Crafting transferable adversarial point clouds via adversarial noise factorization,'' \emph{TBD}, 2024.

\bibitem{Tang-2023-ManifoldAttack}
K.~Tang, J.~Wu, W.~Peng, Y.~Shi, P.~Song, Z.~Gu, Z.~Tian, and W.~Wang, ``Deep manifold attack on point clouds via parameter plane stretching,'' in \emph{AAAI}, vol.~37, no.~2, 2023, pp. 2420--2428.

\bibitem{lou2024hide}
T.~Lou, X.~Jia, J.~Gu, L.~Liu, S.~Liang, B.~He, and X.~Cao, ``Hide in thicket: Generating imperceptible and rational adversarial perturbations on 3d point clouds,'' in \emph{CVPR}, 2024, pp. 24\,326--24\,335.

\bibitem{Qi-2017-Pointnet}
C.~R. Qi, H.~Su, K.~Mo, and L.~J. Guibas, ``Pointnet: Deep learning on point sets for 3d classification and segmentation,'' in \emph{CVPR}, 2017, pp. 652--660.

\bibitem{wu20153d}
Z.~Wu, S.~Song, A.~Khosla, F.~Yu, L.~Zhang, X.~Tang, and J.~Xiao, ``3d shapenets: A deep representation for volumetric shapes,'' in \emph{CVPR}, 2015, pp. 1912--1920.

\bibitem{chang2015shapenet}
A.~X. Chang, T.~Funkhouser, L.~Guibas, P.~Hanrahan, Q.~Huang, Z.~Li, S.~Savarese, M.~Savva, S.~Song, H.~Su \emph{et~al.}, ``Shapenet: An information-rich 3d model repository,'' \emph{arXiv preprint arXiv:1512.03012}, 2015.

\bibitem{uy-scanobjectnn-iccv19}
M.~A. Uy, Q.-H. Pham, B.-S. Hua, D.~T. Nguyen, and S.-K. Yeung, ``Revisiting point cloud classification: A new benchmark dataset and classification model on real-world data,'' in \emph{ICCV}, 2019.

\bibitem{chakraborty2021survey}
A.~Chakraborty, M.~Alam, V.~Dey, A.~Chattopadhyay, and D.~Mukhopadhyay, ``A survey on adversarial attacks and defences,'' \emph{CAAI Transactions on Intelligence Technology}, vol.~6, no.~1, pp. 25--45, 2021.

\bibitem{ren2020adversarial}
K.~Ren, T.~Zheng, Z.~Qin, and X.~Liu, ``Adversarial attacks and defenses in deep learning,'' \emph{Engineering}, vol.~6, no.~3, pp. 346--360, 2020.

\bibitem{wei2024physical}
H.~Wei, H.~Tang, X.~Jia, Z.~Wang, H.~Yu, Z.~Li, S.~Satoh, L.~Van~Gool, and Z.~Wang, ``Physical adversarial attack meets computer vision: A decade survey,'' \emph{TPAMI}, 2024.

\bibitem{Zheng-2019-PCDSaliency}
T.~Zheng, C.~Chen, J.~Yuan, B.~Li, and K.~Ren, ``Pointcloud saliency maps,'' in \emph{ICCV}, 2019, pp. 1598--1606.

\bibitem{Yang-2019-AdvAttackAndDefense}
J.~Yang, Q.~Zhang, R.~Fang, B.~Ni, J.~Liu, and Q.~Tian, ``Adversarial attack and defense on point sets,'' \emph{arXiv preprint arXiv:1902.10899}, 2019.

\bibitem{Wicker-2019-IterSaliencyOcc}
M.~Wicker and M.~Kwiatkowska, ``Robustness of 3d deep learning in an adversarial setting,'' in \emph{CVPR}, 2019, pp. 11\,767--11\,775.

\bibitem{zhang-2021-TopologyDestructionNetwork}
J.~Zhang, C.~Jiang, X.~Wang, and M.~Cai, ``Td-net: Topology destruction network for generating adversarial point cloud,'' in \emph{ICIP}, 2021, pp. 3098--3102.

\bibitem{zhao2020isometry}
Y.~Zhao, Y.~Wu, C.~Chen, and A.~Lim, ``On isometry robustness of deep 3d point cloud models under adversarial attacks,'' in \emph{CVPR}, 2020, pp. 1201--1210.

\bibitem{Kim-2021-minimalAdv}
J.~Kim, B.-S. Hua, T.~Nguyen, and S.-K. Yeung, ``Minimal adversarial examples for deep learning on 3d point clouds,'' in \emph{ICCV}, 2021, pp. 7797--7806.

\bibitem{yang2025hiding}
M.~Yang, D.~Liu, K.~Tang, P.~Zhou, L.~Chen, and J.~Chen, ``Hiding imperceptible noise in curvature-aware patches for 3d point cloud attack,'' in \emph{ECCV}, 2024, pp. 431--448.

\bibitem{Carlini-2017-cw}
N.~Carlini and D.~Wagner, ``Towards evaluating the robustness of neural networks,'' in \emph{S\&P}, 2017, pp. 39--57.

\bibitem{Goodfellow-2014-FGSM}
I.~J. Goodfellow, J.~Shlens, and C.~Szegedy, ``Explaining and harnessing adversarial examples,'' in \emph{ICLR}, 2015.

\bibitem{wen2020geometry}
Y.~Wen, J.~Lin, K.~Chen, C.~P. Chen, and K.~Jia, ``Geometry-aware generation of adversarial point clouds,'' \emph{IEEE TPAMI}, vol.~44, no.~6, pp. 2984--2999, 2022.

\bibitem{huang2022shape}
Q.~Huang, X.~Dong, D.~Chen, H.~Zhou, W.~Zhang, and N.~Yu, ``Shape-invariant 3d adversarial point clouds,'' in \emph{CVPR}, 2022, pp. 15\,335--15\,344.

\bibitem{Tang-NTAttack}
K.~Tang, Y.~Shi, T.~Lou, W.~Peng, X.~He, P.~Zhu, Z.~Gu, and Z.~Tian, ``Rethinking perturbation directions for imperceptible adversarial attacks on point clouds,'' \emph{IEEE Internet of Things Journal}, vol.~10, no.~6, pp. 5158--5169, 2023.

\bibitem{tang2024manifoldConstraints}
K.~Tang, X.~He, W.~Peng, J.~Wu, Y.~Shi, D.~Liu, P.~Zhou, W.~Wang, and Z.~Tian, ``Manifold constraints for imperceptible adversarial attacks on point clouds,'' in \emph{AAAI}, vol.~38, no.~6, 2024, pp. 5127--5135.

\bibitem{tang2024flat}
K.~Tang, L.~Huang, W.~Peng, D.~Liu, X.~Wang, Y.~Ma, L.~Liu, and Z.~Tian, ``Flat: Flux-aware imperceptible adversarial attacks on 3d point clouds,'' in \emph{ECCV}.\hskip 1em plus 0.5em minus 0.4em\relax Springer, 2024, pp. 198--215.

\bibitem{tang2024symattack}
K.~Tang, Z.~Wang, W.~Peng, L.~Huang, L.~Wang, P.~Zhu, W.~Wang, and Z.~Tian, ``Symattack: Symmetry-aware imperceptible adversarial attacks on 3d point clouds,'' in \emph{MM}, 2024, pp. 3131--3140.

\bibitem{Zhou-2020-LGGAN}
H.~Zhou, D.~Chen, J.~Liao, K.~Chen, X.~Dong, K.~Liu, W.~Zhang, G.~Hua, and N.~Yu, ``Lg-gan: Label guided adversarial network for flexible targeted attack of point cloud based deep networks,'' in \emph{CVPR}, 2020, pp. 10\,356--10\,365.

\bibitem{Lee-2020-Shapeadv}
K.~Lee, Z.~Chen, X.~Yan, R.~Urtasun, and E.~Yumer, ``Shapeadv: Generating shape-aware adversarial 3d point clouds,'' \emph{arXiv preprint arXiv:2005.11626}, 2020.

\bibitem{tsai2020robust}
T.~Tsai, K.~Yang, T.-Y. Ho, and Y.~Jin, ``Robust adversarial objects against deep learning models,'' in \emph{AAAI}, vol.~34, no.~01, 2020, pp. 954--962.

\bibitem{zhang20233d}
J.~Zhang, L.~Chen, B.~Liu, B.~Ouyang, Q.~Xie, J.~Zhu, W.~Li, and Y.~Meng, ``3d adversarial attacks beyond point cloud,'' \emph{Information Sciences}, vol. 633, pp. 491--503, 2023.

\bibitem{dong2022isometric}
Y.~Dong, J.~Zhu, X.-S. Gao \emph{et~al.}, ``Isometric 3d adversarial examples in the physical world,'' in \emph{NeurIPS}, vol.~35, 2022, pp. 19\,716--19\,731.

\bibitem{Maturana-2015-voxnet}
D.~Maturana and S.~Scherer, ``Voxnet: A 3d convolutional neural network for real-time object recognition,'' in \emph{IROS}, 2015, pp. 922--928.

\bibitem{Qi-2017-Pointnet++}
C.~R. Qi, L.~Yi, H.~Su, and L.~J. Guibas, ``Pointnet++ deep hierarchical feature learning on point sets in a metric space,'' in \emph{NeurIPS}, 2017, pp. 5105--5114.

\bibitem{PointMLP}
X.~Ma, C.~Qin, H.~You, H.~Ran, and Y.~Fu, ``Rethinking network design and local geometry in point cloud: A simple residual mlp framework,'' in \emph{ICLR}, 2022.

\bibitem{Wu-2019-Pointconv}
W.~Wu, Z.~Qi, and L.~Fuxin, ``Pointconv: Deep convolutional networks on 3d point clouds,'' in \emph{CVPR}, 2019, pp. 9621--9630.

\bibitem{Thomas-2019-KPConv}
H.~Thomas, C.~R. Qi, J.-E. Deschaud, B.~Marcotegui, F.~Goulette, and L.~J. Guibas, ``Kpconv: Flexible and deformable convolution for point clouds,'' in \emph{ICCV}, 2019, pp. 6411--6420.

\bibitem{Xu-2021-PAConv}
M.~Xu, R.~Ding, H.~Zhao, and X.~Qi, ``Paconv: Position adaptive convolution with dynamic kernel assembling on point clouds,'' \emph{CVPR}, 2021.

\bibitem{Li-2018-PointCNN}
Y.~Li, R.~Bu, M.~Sun, W.~Wu, X.~Di, and B.~Chen, ``Pointcnn: Convolution on $\chi$-transformed points,'' in \emph{NeurIPS}, 2018, pp. 820--830.

\bibitem{Wang-2019-DGCNN}
Y.~Wang, Y.~Sun, Z.~Liu, S.~E. Sarma, M.~M. Bronstein, and J.~M. Solomon, ``Dynamic graph cnn for learning on point clouds,'' \emph{TOG}, vol.~38, no.~5, pp. 1--12, 2019.

\bibitem{Zhao-2019-Pointweb}
H.~Zhao, L.~Jiang, C.-W. Fu, and J.~Jia, ``Pointweb: Enhancing local neighborhood features for point cloud processing,'' in \emph{CVPR}, 2019, pp. 5565--5573.

\bibitem{Shi-2020-PointGNN}
W.~Shi and R.~Rajkumar, ``Point-gnn: Graph neural network for 3d object detection in a point cloud,'' in \emph{CVPR}, 2020, pp. 1711--1719.

\bibitem{chen2022ddgcn}
L.~Chen and Q.~Zhang, ``Ddgcn: graph convolution network based on direction and distance for point cloud learning,'' \emph{The Visual Computer}, vol.~39, no.~3, pp. 863--873, 2023.

\bibitem{zhao2021PT}
H.~Zhao, L.~Jiang, J.~Jia, P.~H. Torr, and V.~Koltun, ``Point transformer,'' in \emph{ICCV}, 2021, pp. 16\,259--16\,268.

\bibitem{guo2021pct}
M.-H. Guo, J.-X. Cai, Z.-N. Liu, T.-J. Mu, R.~R. Martin, and S.-M. Hu, ``Pct: Point cloud transformer,'' \emph{Computational Visual Media}, vol.~7, pp. 187--199, 2021.

\bibitem{wu2022PT2}
X.~Wu, Y.~Lao, L.~Jiang, X.~Liu, and H.~Zhao, ``Point transformer v2: Grouped vector attention and partition-based pooling,'' in \emph{NeurIPS}, vol.~35, 2022, pp. 33\,330--33\,342.

\bibitem{wu2024pT3}
X.~Wu, L.~Jiang, P.-S. Wang, Z.~Liu, X.~Liu, Y.~Qiao, W.~Ouyang, T.~He, and H.~Zhao, ``Point transformer v3: Simpler faster stronger,'' in \emph{CVPR}, 2024, pp. 4840--4851.

\bibitem{liang2024pointmamba}
D.~Liang, X.~Zhou, W.~Xu, X.~Zhu, Z.~Zou, X.~Ye, X.~Tan, and X.~Bai, ``Pointmamba: A simple state space model for point cloud analysis,'' in \emph{NeurIPS}, 2024.

\bibitem{ioannidou2017deep}
A.~Ioannidou, E.~Chatzilari, S.~Nikolopoulos, and I.~Kompatsiaris, ``Deep learning advances in computer vision with 3d data: A survey,'' \emph{ACM computing surveys (CSUR)}, vol.~50, no.~2, pp. 1--38, 2017.

\bibitem{anguelov2005scape}
D.~Anguelov, P.~Srinivasan, D.~Koller, S.~Thrun, J.~Rodgers, and J.~Davis, ``Scape: Shape completion and animation of people,'' \emph{TOG}, vol.~24, no.~3, pp. 408--416, 2005.

\bibitem{li2022diffcloth}
Y.~Li, T.~Du, K.~Wu, J.~Xu, and W.~Matusik, ``Diffcloth: Differentiable cloth simulation with dry frictional contact,'' \emph{TOG}, vol.~42, no.~1, pp. 1--20, 2022.

\bibitem{chen2011lattice}
C.-H. Chen, I.-C. Lin, M.-H. Tsai, and P.-H. Lu, ``Lattice-based skinning and deformation for real-time skeleton-driven animation,'' in \emph{International Conference on Computer-Aided Design and Computer Graphics}, 2011, pp. 306--312.

\bibitem{capell2002interactive}
S.~Capell, S.~Green, B.~Curless, T.~Duchamp, and Z.~Popovi{\'c}, ``Interactive skeleton-driven dynamic deformations,'' \emph{TOG}, vol.~21, no.~3, pp. 586--593, 2002.

\bibitem{yoshizawa2007skeleton}
S.~Yoshizawa, A.~Belyaev, and H.-P. Seidel, ``Skeleton-based variational mesh deformations,'' \emph{Computer Graphics Forum}, vol.~26, no.~3, pp. 255--264, 2007.

\bibitem{nieto2012cage}
J.~R. Nieto and A.~Sus{\'\i}n, ``Cage based deformations: a survey,'' in \emph{Deformation Models: Tracking, Animation and Applications}.\hskip 1em plus 0.5em minus 0.4em\relax Springer, 2012, pp. 75--99.

\bibitem{garcia2013cages}
F.~G. Garc{\'\i}a, T.~Paradinas, N.~Coll, and G.~Patow, ``Cages: a multilevel, multi-cage-based system for mesh deformation,'' \emph{TOG}, vol.~32, no.~3, pp. 1--13, 2013.

\bibitem{stroter2024-CageSurvey}
D.~Str{\"o}ter, J.~Thiery, K.~Hormann, J.~Chen, Q.~Chang, S.~Besler, J.~Mueller-Roemer, T.~Boubekeur, A.~Stork, and D.~Fellner, ``A survey on cage-based deformation of 3d models,'' \emph{Computer Graphics Forum}, vol.~43, no.~2, 2024.

\bibitem{ju2023mean}
T.~Ju, S.~Schaefer, and J.~Warren, ``Mean value coordinates for closed triangular meshes,'' in \emph{Seminal Graphics Papers: Pushing the Boundaries, Volume 2}, 2023, pp. 223--228.

\bibitem{Pytorch}
A.~Paszke, S.~Gross, F.~Massa, A.~Lerer, J.~Bradbury, G.~Chanan, T.~Killeen, Z.~Lin, N.~Gimelshein, L.~Antiga, A.~Desmaison, A.~K\"{o}pf, E.~Yang, Z.~DeVito, M.~Raison, A.~Tejani, S.~Chilamkurthy, B.~Steiner, L.~Fang, J.~Bai, and S.~Chintala, ``Pytorch: An imperative style, high-performance deep learning library,'' in \emph{NeurIPS}, 2019, pp. 8026--8037.

\bibitem{PT}
H.~Zhao, L.~Jiang, J.~Jia, P.~H. Torr, and V.~Koltun, ``Point transformer,'' in \emph{ICCV}, 2021, pp. 16\,259--16\,268.

\bibitem{Dong-2020-FGM_PGD_IMPLEMENT}
X.~Dong, D.~Chen, H.~Zhou, G.~Hua, W.~Zhang, and N.~Yu, ``Self-robust 3d point recognition via gather-vector guidance,'' in \emph{CVPR}, 2020, pp. 11\,513--11\,521.

\bibitem{Zhou-2019-dup}
H.~Zhou, K.~Chen, W.~Zhang, H.~Fang, W.~Zhou, and N.~Yu, ``Dup-net: Denoiser and upsampler network for 3d adversarial point clouds defense,'' in \emph{ICCV}, 2019, pp. 1961--1970.

\bibitem{Wu-2020-IFDefense}
Z.~Wu, Y.~Duan, H.~Wang, Q.~Fan, and L.~J. Guibas, ``If-defense: 3d adversarial point cloud defense via implicit function based restoration,'' \emph{arXiv preprint arXiv:2010.05272}, 2020.

\bibitem{li2019pu}
R.~Li, X.~Li, C.-W. Fu, D.~Cohen-Or, and P.-A. Heng, ``Pu-gan: a point cloud upsampling adversarial network,'' in \emph{ICCV}, 2019, pp. 7203--7212.

\end{thebibliography}
}

\end{document}